\documentclass{article}

\usepackage[preprint,nonatbib]{neurips_2026}

\usepackage[utf8]{inputenc}
\usepackage[T1]{fontenc}
\usepackage{hyperref}
\usepackage{url}
\usepackage{booktabs}
\usepackage{amsfonts}
\usepackage{amsmath}
\usepackage{amssymb}
\usepackage{nicefrac}
\usepackage{microtype}
\usepackage[table]{xcolor}
\usepackage{graphicx}
\usepackage{multirow}
\usepackage{enumitem}
\usepackage{makecell}
\usepackage{arydshln}
\usepackage{colortbl}
\usepackage{float}

\newcommand{\best}[1]{\cellcolor{green!15}#1}                                                                                                                                                                                                                       
\newcommand{\second}[1]{\cellcolor{yellow!30}#1}
\newcommand{\third}[1]{\cellcolor{red!15}#1}  
\newcommand{\besttext}[1]{\colorbox{green!15}{#1}}   
\newcommand{\secondtext}[1]{\colorbox{yellow!30}{#1}}                                                                                                                                       
\newcommand{\thirdtext}[1]{\colorbox{red!15}{#1}}
\newcommand{\midruledash}{\noalign{\vskip 2pt} \hdashline \noalign{\vskip 3pt}} 
\newcommand{\qaf}{\ensuremath{Q^{af}}}

\title{Mitigating Content Shift and Hallucination in GenAI Image Editing via Structural Refinement}

\author{%
  Luxi Zhao \quad Michael S. Brown\\
  Department of Electrical Engineering \& Computer Science\\
  York University\\
  \texttt{luxizhao@yorku.ca \quad mbrown@eecs.yorku.ca}
}

\begin{document}

\maketitle

\begin{abstract}
Generative AI (GenAI) image editors, such as Nano Banana, produce visually compelling results for retouching tasks, enabling non-experts to edit images through text prompts alone. However, the generative nature of these models often introduces spatial misalignment, texture distortion, and content hallucination, all of which are detrimental to downstream workflows that require pixel-level fidelity. We identify a problem setting we call \emph{structure-preserving GenAI fusion} for black-box GenAI image retouching: retain the perceptual enhancements of a GenAI output while enforcing structural faithfulness to the original input image. To address this problem, we propose a post-processing framework that fuses an input image with its GenAI-enhanced counterpart by first establishing coarse spatial and photometric correspondences, then performing a fusion stage that transfers desired enhancements while suppressing hallucinated content. In the absence of direct prior work in this setting, we evaluate our framework against representative methods from photorealistic style transfer and image fusion. Our experiments demonstrate that our method better preserves aesthetic quality while maintaining pixel-level structural consistency and the input resolution.
\end{abstract}

\section{Introduction}
\begin{figure}[t]
    \centering
    \includegraphics[width=1.0\linewidth]{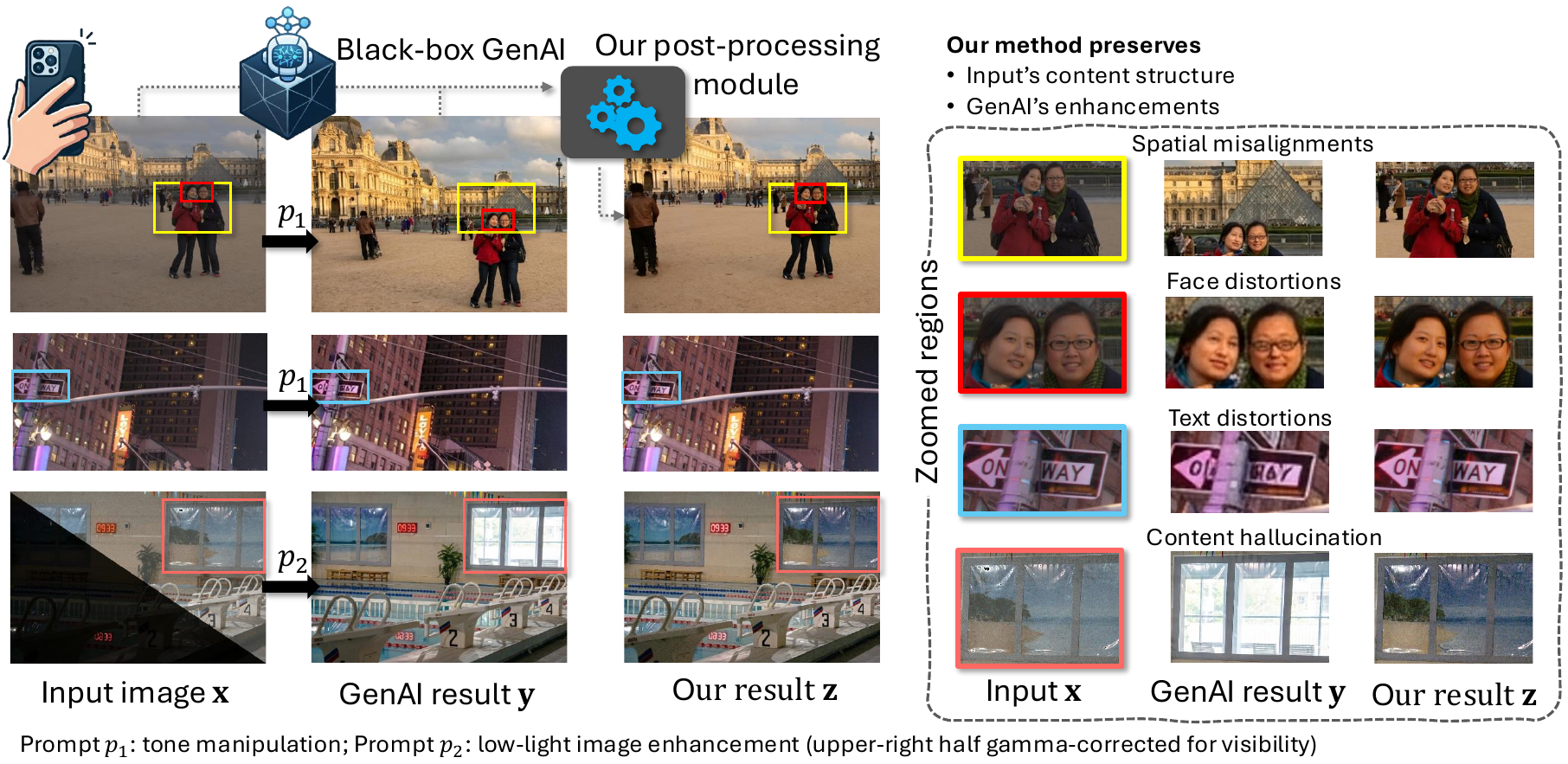}
\vspace{-5mm}
\caption{
  Black-box GenAI image enhancement produces visually compelling outputs $\mathbf{y}$ from input $\mathbf{x}$ 
  (here, via prompts $p_1$ for tone manipulation and $p_2$ for low-light enhancement), 
  but introduces spatial misalignment, texture distortion (faces, text), content hallucination, 
  and resolution mismatch with respect to $\mathbf{x}$. 
  We propose a post-processing method that fuses $\mathbf{x}$ and $\mathbf{y}$ into a result $\mathbf{z}$ that inherits the photometric enhancements of $\mathbf{y}$ while preserving the structure and resolution of $\mathbf{x}$.}
    \label{fig:teaser}
    \vspace{-5mm}
\end{figure}

In recent years, generative AI (GenAI) models have made significant progress in prompt-based image editing.
Their capacity to learn powerful priors from large-scale data allows them to generalize across a wide range of editing tasks, including image retouching.
By entering a text-based prompt, non-expert users can generate enhanced images without learning professional photo-editing tools such as Lightroom or Darktable.

Recent work~\cite{lowlevelbanana} evaluates the zero-shot performance of Nano Banana Pro (NB Pro) across various low-level vision tasks, including tone manipulation, low-light enhancement, denoising, and super-resolution.
While NB Pro excels in perceptual quality, it sacrifices structural fidelity to the original image. 
This makes it unsuitable for downstream workflows that require pixel-level authenticity, such as integration into a professional image-editing pipeline.

In general, we observe that black-box GenAI models exhibit four core failure modes in geometry and content preservation:
(i) \emph{spatial misalignment}, where the GenAI output exhibits unpredictable spatial shifts or warps;
(ii) \emph{texture distortion}, where the GenAI model distorts fine details, especially in perceptually critical regions such as faces and text;
(iii) \emph{content hallucination}, where the GenAI model introduces non-existing content, such as clouds in the sky or extra people in the background;
(iv) \emph{resolution mismatch}, where the GenAI output is lower in spatial resolution than the full-resolution camera input, leading to loss of high-frequency content.
See Figure~\ref{fig:teaser} for examples.
While such failures are often perceptible to a human observer, quantifying them at scale remains difficult.
No-reference metrics assess perceptual quality but do not capture faithfulness to the input, while reference-based metrics capture only selected aspects of output quality in this setting.
As we observe in our experiments, different metrics can yield differing assessments of the same output, suggesting that no single metric, in isolation, fully captures this class of problems.

We focus on enhancement-based tasks such as tone manipulation and low-light image enhancement, where the desired edits are primarily color and tonal adjustments, and the original camera image therefore serves as a natural reference for scene structure.
This setting allows us to leverage the accessibility and powerful priors of frontier black-box GenAI models while enforcing structural fidelity to the input.
We do not target tasks such as super-resolution or deblurring, where the notion of ground-truth scene structure is itself ill-posed.
Although we adopt the camera image as our primary use case---reflecting the common scenario of a user capturing a photo on a phone and editing it with a GenAI tool---our framework applies in principle to any input image used as a structural reference.
To our knowledge, no prior work systematically addresses the challenges of black-box GenAI image enhancement; the two most closely related bodies of work are photorealistic style transfer and multimodal image fusion.

Photorealistic style transfer (PST) methods~\cite{salut,photowct2,wct2} transfer the photofinishing style of a style image onto a content image while preserving the latter's structure.
They handle high-resolution images and tolerate spatial misalignment, but their stylization is often global and lacks strict semantic consistency, which can lead to incorrect stylization of local regions.
In addition, PST methods often target clean, high-quality images~\cite{dpst,flickr2k}; they cannot suppress noise in the input image or exploit the already-denoised GenAI output.

Multimodal image fusion methods~\cite{cunet,ddfm,rffusion,cddfuse,swinfusion} merge complementary information from two or more source images of different modalities into a single image with improved scene representation.
Common applications include infrared--visible fusion and multi-exposure fusion.
These methods typically employ self-supervised objectives that encourage the fused image to inherit structural characteristics from both sources.
In our setting, however, the GenAI output contains hallucinations and cannot be treated as a reliable source of structural information.

We address these challenges by proposing a generic camera--GenAI image fusion framework that preserves the desirable improvements of the GenAI output---such as photofinishing effects and detail enhancements---while enforcing geometric and spatial-resolution consistency with the camera image.
The framework follows a three-stage pipeline of flow-based spatial alignment, global color and tone transfer, and an interpretable fusion module based on sparse coding.
We demonstrate our method on tone manipulation and low-light enhancement, where it shows strong performance in both content preservation and style transfer compared to PST and image fusion baselines.

In summary, our contributions are as follows:
\begin{itemize}[leftmargin=10pt, noitemsep, topsep=0pt]
\item We propose a post-processing framework that fuses a GenAI-enhanced image with the original camera image in three stages---spatial alignment, color and tonal alignment, and an interpretable fusion module based on sparse coding---to address spatial misalignment and content hallucination in black-box GenAI image enhancement.
\item We introduce a supervised training pipeline with synthetic data that is generic across enhancement-based settings, and evaluate the framework on tone manipulation and low-light enhancement.
\end{itemize}
  
\section{Related Work}

No prior work directly addresses hallucination correction and misalignment handling in black-box GenAI image enhancement.
We identify two closely related task settings: photorealistic style transfer and image fusion.
Photorealistic style transfer (Sec.~\ref{sec:pst}) transfers color and tonal characteristics while preserving structure, and image fusion (Sec.~\ref{sec:fusion}) combines complementary information from multiple source images of the same scene.
Since our method builds on convolutional sparse coding, a method commonly used for interpretable image fusion, we further review CSC-based fusion in Sec.~\ref{sec:csc}.
Together, these areas inform our framework, but none directly addresses our problem setting.

\subsection{Photorealistic style transfer}
\label{sec:pst}
Photorealistic style transfer (PST) aims to transfer a style image's color and tonal characteristics to a content image while strictly preserving the content image's spatial structure.
This contrasts with artistic style transfer~\cite{gatys2016image,arbitrary_adain}, which allows significant texture and structural distortions.

A foundational line of work~\cite{photowct,wct2,photowct2} builds on the Whitening and Coloring Transform (WCT)~\cite{wct}, which aligns the correlation matrices of content feature maps---typically extracted using a pretrained VGG~\cite{vgg19} network---with those of the style feature maps.
The aligned features are decoded into the output image via a pretrained decoder, with skip connections from the encoder to preserve detail.

Preset-based methods~\cite{deeppreset,neuralpreset} predict global color styles from the style image and apply them to the content image in feature space.
NeuralPreset~\cite{neuralpreset}, for example, learns a pair of normalizing and stylizing matrices that adjust each pixel's color independently, 
irrespective of the pixel's spatial location and neighboring pixels.
This is effective at preserving the content image's structure, but is limited in capturing local style variations.

SA-LUT~\cite{salut} learns a spatially adaptive 4D look-up table for photorealistic style transfer.
Its 4th dimension consists of two context bins, indexed by a spatially-varying context map that identifies corresponding regions between the content and style images.

Two structural assumptions of PST methods limit their applicability to our setting. 
First, PST is designed for loosely aligned or unrelated content--style pairs, so its underlying mechanisms---global feature-statistic matching in WCT-based methods, global color predictors in preset-based methods, and SA-LUT's spatially adaptive but low-dimensional context modulation---do not fully exploit dense pixel-level correspondence.
In our setting, the camera image and GenAI output are densely aligned (after our spatial alignment stage), and these mechanisms cannot fully exploit that correspondence.
Second, PST assumes a clean reference image; the methods do not exploit detail enhancements in the reference, such as denoising, that are characteristic of GenAI outputs.

\subsection{Multimodal image fusion}
\label{sec:fusion}
Multimodal image fusion combines multiple images from the same scene into a single image that contains complementary information from each.
Existing approaches span GAN-based~\cite{ma2020infrared,ma2019fusiongan,ganmcc}, transformer-based~\cite{swinfusion,cddfuse}, autoencoder-based~\cite{li2018densefuse,sdnet}, convolutional sparse coding-based~\cite{cunet,DBLP:journals/tip/GaoDXXD22,DBLP:journals/corr/abs-2005-08448,deepm2cdl}, and diffusion-based methods~\cite{ddfm,cao2024conditional,rffusion}.

A common objective in image fusion is to maximize the information the fused image inherits from each source~\cite{scd}, leading to self-supervised losses that encourage structural similarity to both inputs.
For example, CDDFuse~\cite{cddfuse} and SwinFusion~\cite{swinfusion} use averaged reconstruction losses between the fused image and each source.
Diffusion-based fusion methods also rely on conventional fusion objectives:
DDFM~\cite{ddfm} derives a likelihood rectification step for DDPM sampling from a standard intensity-based fusion loss,
while RFFusion~\cite{rffusion} extends this with rectified flow and a fusion-specific VAE.
Thus, although these methods use stronger generative priors, their fusion guidance remains tied to objectives that encourage retaining information from both sources.

Image fusion objectives assume both source images are reliable structural references.
The GenAI output, however, contains hallucinations and misalignments. Therefore, encouraging the fused image to inherit GenAI-side structure is counterproductive.
Some image-fusion architectures may be reusable, but their training or sampling objectives are not directly applicable to our task.

\subsection{Interpretable fusion via convolutional sparse coding}
\label{sec:csc}
Most PST and image fusion methods~\cite{photowct2,neuralpreset,cddfuse,swinfusion} produce outputs without an explicit account of how each output region relates to the inputs, limiting their interpretability.
Convolutional sparse coding (CSC) methods~\cite{DBLP:journals/tip/GaoDXXD22,cunet} provide a more interpretable alternative by decomposing source images into shared (common) and modality-specific (unique) components.
The unique components localize where the source images diverge, offering a natural representation for analyzing inconsistencies between two images of the same scene.
We build on this property of CSC-based fusion in our framework, as detailed in Sec.~\ref{sec:method}.

\section{Method}
\label{sec:method}
\begin{figure}[t]
    \centering
    \includegraphics[width=1.0\linewidth]{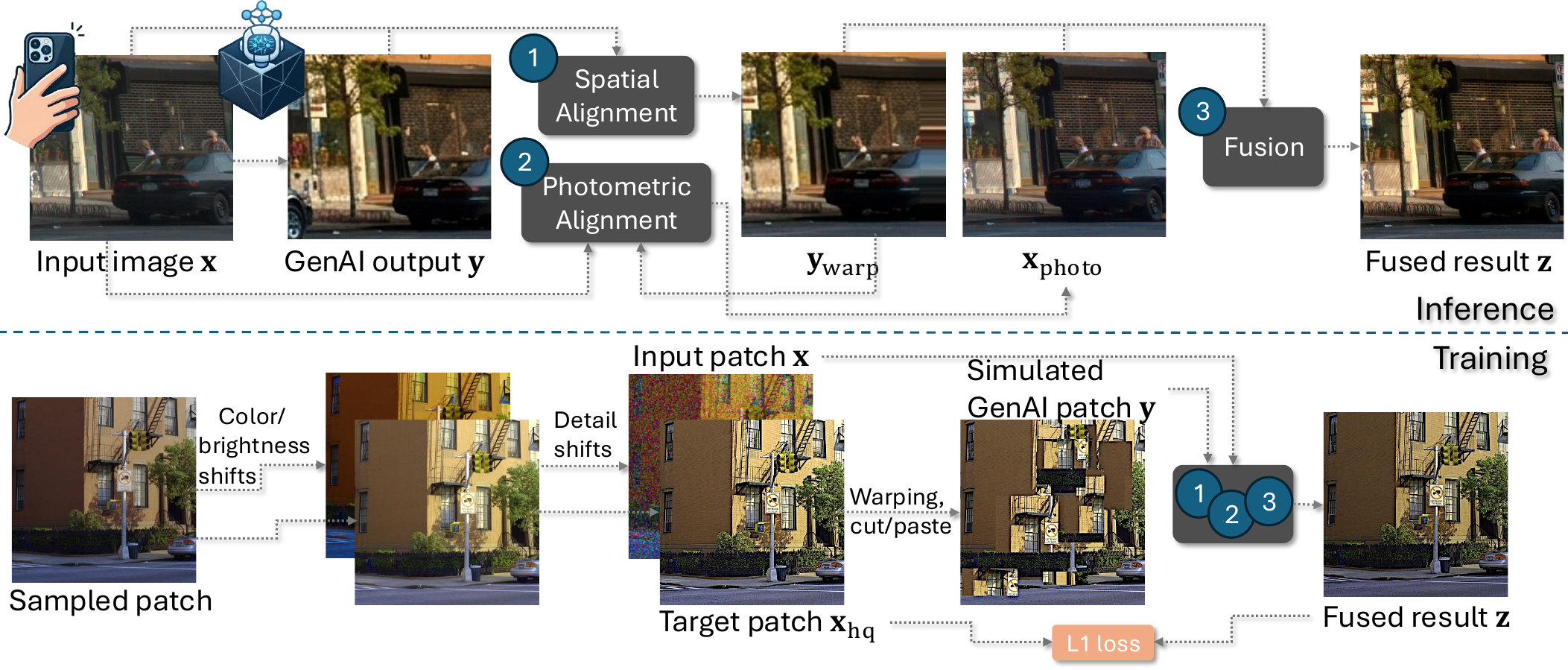}
    \caption{Overview of our method. 
    Top: during inference, 1) optical-flow alignment warps the GenAI output $\mathbf{y}$ to the camera input $\mathbf{x}$; 
    2) a global photometric alignment module (PAM: S-curve, CbCr LUT, and gamma correction) maps $\mathbf{x}$ to $\mathbf{y}_{\text{warp}}$ while avoiding hallucination leakage; 
    and 3) MSCU-Net with a lightweight NAFNet refiner fuses the aligned pair into the output $\mathbf{z}$.
    Bottom: during training, triplets $(\mathbf{x}_{hq}, \mathbf{x}, \mathbf{y})$ are synthesized from clean patches using color/brightness shifts, detail shifts, and synthetic hallucination.}
    \label{fig:method}
\end{figure}
\vspace{-1mm}

In this section, we introduce our post-processing framework for camera--GenAI fusion.
Our fusion framework consists of the following main components: 
spatial and photometric alignment (Sec.~\ref{subsec:align}),
interpretable fusion model (Sec.~\ref{subsec:fusion}), and a
training pipeline based on synthetic data (Sec.~\ref{subsec:data_pipeline}).
Figure~\ref{fig:method} shows an overview of our method. 

\textbf{Framework overview.} 
We adopt the following notation throughout the paper. 
Given an input image $\mathbf{x} \in \mathbb{R}^{H \times W \times 3}$ and its GenAI-enhanced version $\mathbf{y} \in \mathbb{R}^{h \times w \times 3}$, 
we aim to obtain a fused image $\mathbf{z} \in \mathbb{R}^{H \times W \times 3}$ 
that combines the content structure of $\mathbf{x}$ with the aesthetic quality of $\mathbf{y}$. 
Note that $\mathbf{y}$ may have lower spatial resolution than $\mathbf{x}$ due to GenAI model output constraints.

Overall, our proposed framework can be summarized as follows:
\begin{equation}
\mathbf{y}_{\text{warp}} = F(\mathbf{x}, \mathbf{y}), \quad 
\mathbf{x}_{\text{photo}} = P(\mathbf{x}, \mathbf{y}_{\text{warp}}), \quad 
\mathbf{z} = R(\mathbf{x}_{\text{photo}}, \mathbf{y}_{\text{warp}})
\end{equation}
where $F$, $P$, and $R$ are the spatial alignment module, the photometric alignment module, and the fusion module, respectively.

\subsection{Spatial and Photometric Alignment}
\label{subsec:align}
\textbf{Spatial alignment.} 
We use a pretrained optical flow model to perform spatial alignment: $\mathbf{y}_{\text{warp}} = F(\mathbf{x}, \mathbf{y})$, where $F$ is the flow model.
While optical flow is effective at addressing geometric misalignment, it cannot address texture or content hallucination.
In addition, since GenAI models can produce outputs with changes in field of view, leading to missing or hallucinated content at the image border,
optical flow cannot recover content that is present in the input but missing in the GenAI output (for example, the bottom ground region in Figure~\ref{fig:teaser} row 1).
Sec.~\ref{subsec:fusion} addresses these issues.

\textbf{Photometric alignment.} 
The photometric alignment module (PAM) aims to match the overall color and contrast of the input $\mathbf{x}$ to those of the flow-corrected GenAI reference $\mathbf{y}_{\text{warp}}$ as closely as possible.
Crucially, this stage uses global image processing operations only. 
This avoids the risk of local operations, such as local tone mapping, 
carrying over the color components of hallucinated content from the reference. 
Inspired by the global components of the modular ISP in~\cite{afifi2025modular}, our pipeline 
consists of a generalized global S-curve for contrast adjustment, a CbCr 2D look-up table (LUT), and a gamma correction module.
See Sec.~\ref{appendix:photometric_alignment_module} for details.

By applying spatial and photometric alignment, 
we substantially reduce pixel-level discrepancies between $\mathbf{x}$ and $\mathbf{y}$.
This allows the subsequent fusion model to more easily distinguish hallucinated regions in $\mathbf{y}$ 
from genuine color and tonal adjustments.

\subsection{Interpretable Fusion Model}
\label{subsec:fusion}
\subsubsection{Convolutional sparse coding and common--unique decomposition}
Convolutional sparse coding (CSC)~\cite{csc:lucey, csc:heide, csc:deconv} models a signal $\mathbf{x} \in \mathbb{R}^{H \times W \times K}$ as
\begin{equation}
\mathbf{x} = \sum_{m=0}^{M-1} \mathbf{d}_{m} * \mathbf{u}_{m} = \mathbf{d} * \mathbf{u},
\end{equation}
where $*$ is the convolution operator, $\mathbf{d}=\{ \mathbf{d}_{m} \}_{m=0}^{M-1} \in \mathbb{R}^{s \times s \times M \times K}$ is a convolutional dictionary with $M$ input channels and $K$ output channels.
$\mathbf{u} = \{\mathbf{u}_{m} \}_{m=0}^{M-1} \in \mathbb{R}^{H \times W \times M}$ is a set of sparse feature maps. 
Each dictionary atom $\mathbf{d}_m \in \mathbb{R}^{s \times s \times K}$ produces a $K$-channel contribution from the sparse map $\mathbf{u}_m \in \mathbb{R}^{H \times W}$.
Because $\mathbf{u}$ is enforced to be sparse, it typically captures the content structure of $\mathbf{x}$, while the filters $\mathbf{d}$ capture texture characteristics. 

The CSC objective can be optimized via unrolling the Iterative Shrinkage and Thresholding Algorithm (ISTA) into a fixed-depth neural network (convolutional LISTA) \cite{lcsc}, enabling end-to-end training.

Works~\cite{cunet, DBLP:journals/tip/GaoDXXD22} apply this model to multimodal image fusion by decomposing the two input images $\mathbf{x}$ and $\mathbf{y}$ into common and unique components:
\begin{equation}
  \label{eq:decomp}
\mathbf{x} = \mathbf{d}^c\ast\mathbf{c}+\mathbf{d}^u\ast\mathbf{u}, \quad
\mathbf{y} = \mathbf{h}^c\ast\mathbf{c}+\mathbf{h}^v\ast\mathbf{v}, \quad
\mathbf{z} = \mathbf{g}^c\ast\mathbf{c}+\mathbf{g}^u\ast\mathbf{u}. 
\end{equation}

Here, $\mathbf{x}$ is decomposed into a common sparse component $\mathbf{c}$ and a unique component $\mathbf{u}$; 
$\mathbf{y}$ is decomposed into the same common component $\mathbf{c}$ and a unique component $\mathbf{v}$; 
$\mathbf{d}^c, \mathbf{h}^c, \mathbf{d}^u, \mathbf{h}^v$ are learned filters/dictionaries corresponding to the common and unique components;
$\mathbf{g}^c$ and $\mathbf{g}^u$ are dictionaries adapted to the characteristics of the fused output $\mathbf{z}$.
For our task, since the output should rely on the structure of $\mathbf{x}$, 
we omit $\mathbf{v}$, the unique component of $\mathbf{y}$, during the reconstruction of $\mathbf{z}$.

In~\cite{cunet}, the sparse codes are predicted by separate LISTA modules with update step $(j)$:
\begin{align}
&\mathbf{u}^{(j+1)} = S_{\theta_j^u}\!\left( \mathbf{u}^{(j)} - \mathbf{d}_e^u * \mathbf{d}_d^u * \mathbf{u}^{(j)} + \mathbf{d}_e^u * \mathbf{\hat{x}} \right), \quad \mathbf{\hat{x}}=\mathbf{x} - \mathbf{d}^c * \mathbf{c} \label{eq:ista_u}\\
&\mathbf{c}^{(j+1)} = S_{\theta_j^c}\!\left( \mathbf{c}^{(j)} - \mathbf{w}_e^c * \mathbf{w}_d^c * \mathbf{c}^{(j)} + \mathbf{w}_e^c * [\mathbf{\tilde{x}}, \mathbf{\tilde{y}}] \right), \quad \mathbf{\tilde{x}}=\mathbf{x} - \mathbf{d}^u * \mathbf{u}, \mathbf{\tilde{y}}=\mathbf{y} - \mathbf{h}^v * \mathbf{v} \label{eq:ista_c}
\end{align}
where $S_{\theta_j^u}$ is a soft-thresholding operator with learned threshold $\theta_j^u$. 
For prediction of $\mathbf{u}$, $\mathbf{d}_e^u$ and $\mathbf{d}_d^u$ are learnable convolutional filters trained to 
approximate dictionary $\mathbf{d}^u$. Intuitively, $\mathbf{d}^u_d$ decodes a sparse code into the image domain, while $\mathbf{d}^u_e$ encodes an image into a sparse code.
The same structure applies to $\mathbf{v}$. 
For prediction of $\mathbf{c}$, $\mathbf{w}_d^c$ parameterizes $[\mathbf{d}^c, \mathbf{h}^c]$, where $[\cdot]$ denotes channel-wise concatenation.
$\mathbf{\hat{x}}$ and $\mathbf{\hat{y}}$ represent the image-space unique residuals of $\mathbf{x}$ and $\mathbf{y}$, respectively.
$\mathbf{\tilde{x}}$ and $\mathbf{\tilde{y}}$ represent their image-space common residuals.

\subsubsection{Multiscale common--unique decomposition}
While CU-Net~\cite{cunet} is effective for image fusion, we find that its single-scale formulation deviates from the original mathematical model in ways that hurt interpretability of hallucinated regions in the GenAI output.
Specifically, $\mathbf{\hat{x}}$ and $\mathbf{\hat{y}}$ are unavailable to the feed-forward network, because they depend on the unknown code $\mathbf{c}$ (Eq.~\ref{eq:ista_u}), 
which is predicted only after $\mathbf{u}$ and $\mathbf{v}$ have been computed (Eq.~\ref{eq:ista_c}). 
As a result, $\mathbf{x}$ and $\mathbf{y}$ are used as inputs to the unique-code predictors instead of $\mathbf{\hat{x}}$ and $\mathbf{\hat{y}}$.
This deviation from the original formulation can lead to inaccuracies in the sparse code prediction of $\mathbf{u}$ and $\mathbf{v}$, 
making the resulting decomposition less interpretable.

A pyramidal architecture lets us initialize the coarsest scale with $\mathbf{x}$ and $\mathbf{y}$ (Eq.~\ref{xy_init}),
and propagate the predicted common-component residual to form $\mathbf{\hat{x}}$ and $\mathbf{\hat{y}}$ at finer scales (Eq.~\ref{xy_hat}).
The multiscale architecture can also be viewed as iterative refinement of the unique components $\mathbf{\hat{x}}$ and $\mathbf{\hat{y}}$, 
yielding a closer approximation of the underlying mathematical model.
In addition, with a multiscale architecture, we can warm-start finer-scale LISTA iterations with upsampled sparse codes from coarser scales (Eq.~\ref{eq:warm_start}):
\vspace{-1mm}
\begin{align}
&\mathbf{\hat{x}}^{(0)}=\mathbf{x}^{(0)}, \quad 
\mathbf{\hat{y}}^{(0)}=\mathbf{y}^{(0)} \label{xy_init} \\ 
&\mathbf{\hat{x}}^{(s+1)}=\mathbf{x}^{(s+1)} - \uparrow (\mathbf{d}^{c,(s)} * \mathbf{c}^{(s)}), \quad
\mathbf{\hat{y}}^{(s+1)}=\mathbf{y}^{(s+1)} - \uparrow (\mathbf{h}^{c,(s)} * \mathbf{c}^{(s)}) \label{xy_hat} \\ 
&\mathbf{u}^{(0),(s+1)}= \uparrow \mathbf{u}^{(J-1),(s)}, \quad
\mathbf{v}^{(0),(s+1)}= \uparrow \mathbf{v}^{(J-1),(s)}, \quad
\mathbf{c}^{(0),(s+1)}= \uparrow \mathbf{c}^{(J-1),(s)} \label{eq:warm_start}
\end{align}
where $\mathbf{x} \in \mathbb{R}^{H \times W \times C}$, $\mathbf{y} \in \mathbb{R}^{h \times w \times C}$, 
$s$ is the scale index, $\uparrow$ denotes upsampling, and $J$ is the number of LISTA iterations per scale.
Figure~\ref{fig:model} in the Appendix illustrates the model architecture.

\textbf{Training loss.} 
Since the multiscale network propagates residual inputs (e.g. $\mathbf{\hat{x}}$ and $\mathbf{\hat{y}}$) and sparse codes across scales, 
supervising only the finest-scale output provides weak constraints on the coarser-scale decompositions.
We therefore supervise the fused output $\mathbf{z}^{(s)}$ at each scale, which encourages the intermediate decompositions to remain predictive of the target rather than serving only as latent features:
\begin{equation}
  \mathcal{L} =
\|\mathbf{z}^{(2)} - \mathbf{x}_{hq}^{(2)}\|_1
+ \alpha \|\mathbf{z}^{(1)} - \mathbf{x}_{hq}^{(1)}\|_1
+ \alpha \|\mathbf{z}^{(0)} - \mathbf{x}_{hq}^{(0)}\|_1 .
  \label{eq:loss}
\end{equation}
where $\mathbf{z}^{(s)}$ and $\mathbf{x}_{hq}^{(s)}$ are the output and ground truth at scale $s$, with $(2)$ being the finest scale for a three-scale network. $\alpha < 1$ down-weights coarser scales.

\textbf{Residual refinement.} 
While the sparse coding-based decomposition provides interpretability and explicit control over GenAI-unique content, 
we find that a lightweight restoration-style refiner improves perceptual quality.
We therefore finetune the multiscale CU-Net with a small NAFNet~\cite{nafnet} refinement module.
Details are provided in Appendix~\ref{appendix:nafnet_addon}.

\subsection{Data Synthesis Pipeline}
\label{subsec:data_pipeline}
Since there is no ground truth for the fused output $\mathbf{z}$, we train our model with synthetic data. 
Figure~\ref{fig:method} shows an overview of the data generation pipeline.
We construct triplets $(\mathbf{x}_{hq}, \mathbf{x}, \mathbf{y})$, where $\mathbf{x}_{hq}$ is a high-quality clean image, 
$\mathbf{x}$ is a degraded version of $\mathbf{x}_{hq}$, and $\mathbf{y}$ is the pseudo-GenAI output.

We start from a source patch and apply two different settings of random color and contrast jittering 
to create two versions of the same scene with different photometric properties: one used as the input and the other as ground truth.
We then apply random blur and noise to simulate the lower-quality input $\mathbf{x}$, 
and USM sharpening with random strength to augment the ground truth $\mathbf{x}_{hq}$.
The pseudo-GenAI output $\mathbf{y}$ is generated by applying a homography to $\mathbf{x}_{hq}$ to simulate spatial shifts, and randomly cut-pasting patches to 
create spatially-misplaced content. This encourages the model to rely on $\mathbf{x}$ for structural guidance rather than directly copying from $\mathbf{y}$. 

To help the model better recognize GenAI texture distortions, we apply the SDXL ControlNet image-to-image pipeline~\cite{sdxl} to deliberately hallucinate texture in $\mathbf{x}_{hq}$. 
Two ControlNets---a tile ControlNet for low-frequency structure and a Canny ControlNet for edge preservation---guide the generation. 
The denoising strength is randomly sampled to vary the degree of hallucination.

\begin{figure}[H]
    \centering
    \includegraphics[width=0.95\linewidth]{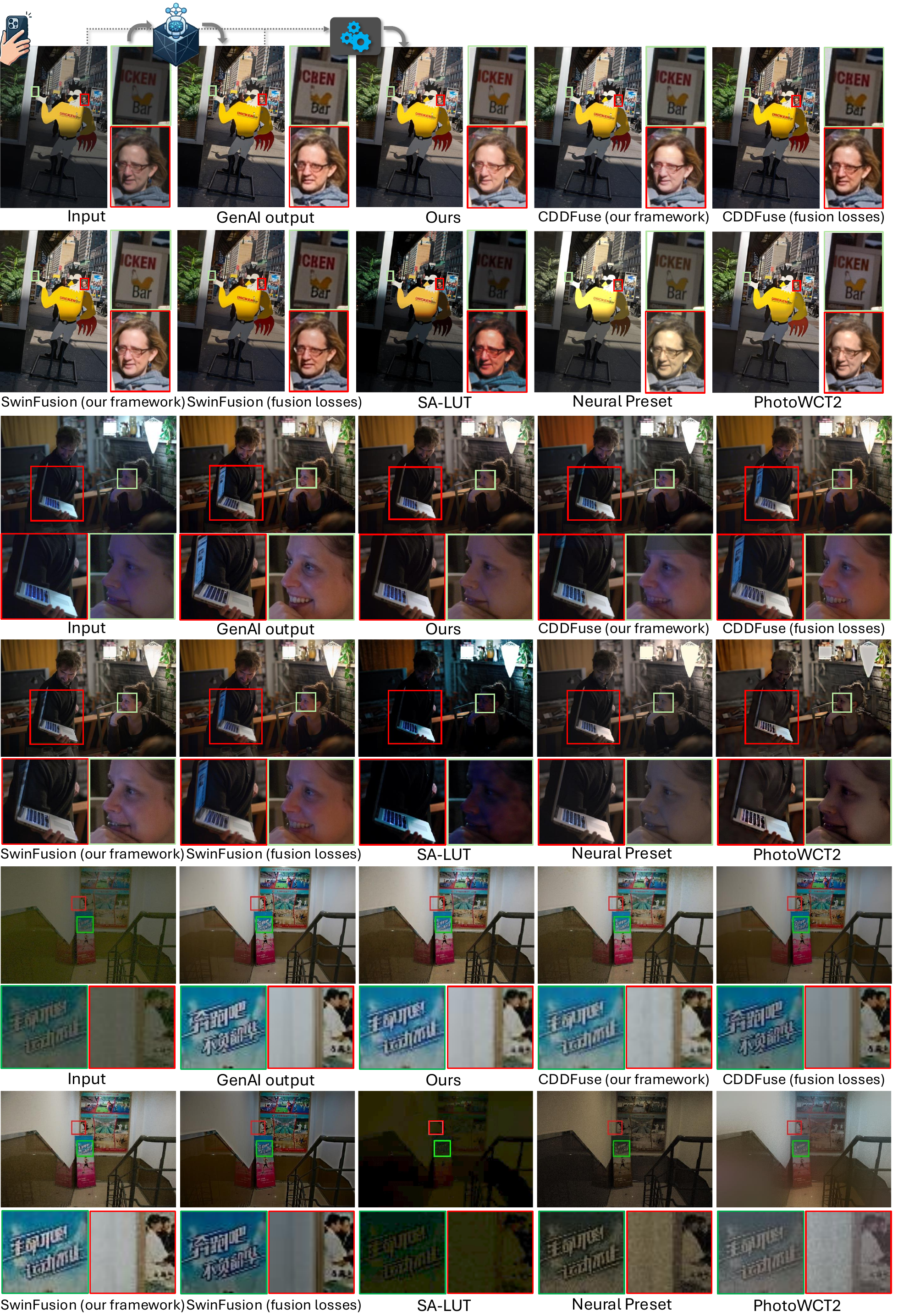}
    \caption{Qualitative comparison on MIT-Adobe FiveK~\cite{ma5k} (top rows, tone manipulation) and LOL v2~\cite{lol_v2} (bottom rows, low-light enhancement). Image fusion baselines trained with their original objectives propagate hallucinated structures from $\mathbf{y}$ (e.g., distorted text and faces); photorealistic style transfer methods preserve structure but miss local stylistic effects under large brightness gaps. Our method preserves the structure of $\mathbf{x}$ while matching the tonal style of $\mathbf{y}$.}
    \label{fig:qualitative_main}
\end{figure}

\section{Experiments}
\label{sec:experiments}
\vspace{-3mm}

\textbf{Datasets.}
We evaluate on two tasks: tone manipulation and low-light enhancement. 
For training, we use the Expert C ground truth from MIT-Adobe FiveK (MA5K)~\cite{ma5k}, with 4000 scenes for training and 500 for validation. We train a single model across both tasks, demonstrating the framework's generality.
We test on MA5K for tone manipulation, and on an aggregate of LOL v1~\cite{lol_v1}, LOL v2-real~\cite{lol_v2}, and SICE~\cite{sice} for low-light enhancement, with equal weighting between LOL and SICE.
Since our framework is most relevant when GenAI outputs exhibit visible misalignment or hallucination, we evaluate on a manually-selected MA5K subset of 285 such scenes.

\textbf{Implementation details.}
We use SKFlow~\cite{skflow} for spatial alignment.
The multiscale CU-Net (MSCU-Net) uses 3 scales with downsampling factor 2 per scale.
We pre-train MSCU-Net with a prepended flow estimator for 400 epochs on MA5K~\cite{ma5k} using Adam~\cite{adam} with learning rate $10^{-5}$, following~\cite{cunet}. 
This is to ensure that the model has the capacity to handle large photometric differences.
In parallel, we train the photometric alignment module (PAM) for 500 epochs with learning rate $10^{-3}$.
We then finetune MSCU-Net jointly with PAM for 50 epochs, and finally train a small NAFNet refiner with frozen MSCU-Net for 200 epochs.

\textbf{Baselines.}
We compare against (1) photorealistic style transfer (PST) methods, including SA-LUT~\cite{salut}, Neural Preset~\cite{neuralpreset}, and PhotoWCT2~\cite{photowct2}; and (2) multimodal image fusion methods, CDDFuse~\cite{cddfuse} and SwinFusion~\cite{swinfusion}.
As PST methods are not dataset-specific, we directly use the pretrained models without re-training. 
For fusion methods, we evaluate two versions: (2.1) trained on our data pipeline with their original fusion losses, and (2.2) embedded in our framework with pre-appended flow and photometric alignment modules.
For (2.2), we use the same training procedure, flow estimator, and PAM as our model for fair comparison.

\textbf{Qualitative Results.}
As shown in Figure~\ref{fig:qualitative_main}, our method preserves input structure while matching the GenAI output's tonal style.
Fusion methods trained with their original losses propagate texture distortions and hallucinations from $\mathbf{y}$, since their objectives encourage inheriting edge information from both sources. When integrated into our framework, the same backbones are better able to reject hallucinations.
PST methods preserve structure faithfully but often fail to match local stylistic effects, especially under large brightness gaps.

\textbf{Metrics.}
We report three metric groups: content fidelity, style similarity, and no-reference IQA. 

For \emph{content fidelity}, we use content similarity~\cite{neuralpreset}, which computes the SSIM between edge maps of $\mathbf{x}$ and $\mathbf{z}$, and \qaf, a component of the classical image fusion metric $Q^{ab/f}$. The latter is computed as $\qaf(i,j)=Q_g^{af}(i,j)Q_\alpha^{af}(i,j)$, where \qaf measures the amount of edge information from source image $a$ preserved in fused image $f$ at location $(i,j)$.
$Q_g^{af}$ and $Q_\alpha^{af}$ are Sobel-based edge-strength and edge-orientation preservation terms.
For \emph{style similarity}, we follow~\cite{w2dist} and use the $W_2$ distance between the global color distributions of $\mathbf{z}$ and the GenAI reference, plus a local $W_2$ averaged over a $6 \times 6$ grid to capture spatially varying style differences.
For \emph{perceptual quality}, we use the no-reference (NR-IQA) metrics MANIQA~\cite{maniqa} and MUSIQ~\cite{musiq}.

\textbf{Quantitative Results.}
As shown in Table~\ref{tab:main-ma5k} and Table~\ref{tab:main-lowlight},
our method achieves a favorable balance across content fidelity, style similarity, and perceptual quality.
Since no single metric captures the combined goal of structural preservation and photometric enhancement, the metrics must be interpreted jointly.
PST methods achieve strong content fidelity but underperform on style and perceptual quality, especially on low-light data.
Fusion baselines trained with their original losses score low on content fidelity, indicating that they inherit structure from the GenAI output $\mathbf{y}$ as well as the input $\mathbf{x}$; their high NR-IQA scores on low-light datasets reflect this propagation of both the desirable perceptual qualities of the GenAI reference $\mathbf{y}$ and its hallucinations (Figure~\ref{fig:qualitative_main}).
When embedded in our framework, these backbones substantially improve on content fidelity and style similarity, confirming that our formulation is backbone-agnostic.
For reference, we also report metrics between $\mathbf{x}$ and $\mathbf{y}$ (resized to input resolution for NR-IQA); the low MUSIQ on MA5K reflects $\mathbf{y}$'s loss of fine detail at lower resolution.

\textbf{Efficiency.}
Fusion baselines have high memory footprints despite modest parameter counts, since they generate full-resolution feature maps during inference; 
this often forces tiled inference, introducing visible seam artifacts (see Figure~\ref{fig:qualitative_main}, row 3, column 4). 
Our method maintains a lower memory footprint with competitive runtime, making it more suitable for full-resolution inputs. See Appendix~\ref{appendix:complexity} for details on inference cost.

\textbf{Framework Analysis.}
Ablations are provided in Appendix~\ref{appendix:ablations}.
Although our framework is compatible with different fusion backbones,
We default to the residual-refined MSCU-Net for its balance of fidelity, style transfer, and efficiency. 
Its built-in interpretability produces a hallucination map as a helpful byproduct,
enabling analysis of hallucination rates across prompts (example shown in~\ref{appendix:prompts}), models, and inputs. 
See Appendix~\ref{appendix:interpretability} for analysis on interpretability.

\vspace{-2mm}
\begin{table}[htbp]
\centering
\caption{Results on MIT-Adobe FiveK~\cite{ma5k}. 
The \besttext{best}, \secondtext{second}, and \thirdtext{third} results are highlighted 
per method group.}
\label{tab:main-ma5k}
\small          
\setlength{\tabcolsep}{3.5pt} 
\begin{tabular}{lcccccc}
\toprule
Experiment & \makecell{\textbf{Content} \\ \textbf{Sim.}$\uparrow$} & \textbf{Qaf}$\uparrow$ & \textbf{W2 Dist.}$\downarrow$ & \makecell{\textbf{W2 Dist.} \\ \textbf{(local)}$\downarrow$} & \textbf{MANIQA}$\uparrow$ & \textbf{MUSIQ}$\uparrow$ \\
\midrule
  Input/GenAI output & 0.503 & 0.177 & 0.205 & 0.201 & 0.293 & 30.61 \\ \midruledash
  SA-LUT~\cite{salut} & \third{0.843} & \third{0.737} & 0.232 & 0.230 & 0.254 & 33.61 \\
  Neural Preset~\cite{neuralpreset} & \best{0.992} & \best{0.929} & 0.094 & 0.103 & \second{0.273} & 34.86 \\
  PhotoWCT2~\cite{photowct2} & 0.787 & 0.632 & \third{0.065} & 0.096 & \best{0.289} & 34.99 \\ \midruledash
  CDDFuse~\cite{cddfuse} (w. fusion losses) & 0.646 & 0.478 & 0.081 & 0.082 & 0.266 & 33.00 \\
  CDDFuse~\cite{cddfuse} (w. our framework) & 0.789 & 0.582 & 0.071 & \third{0.072} & 0.257 & \second{35.85} \\
  SwinFusion~\cite{swinfusion} (w. fusion losses) & 0.625 & 0.428 & 0.100 & 0.101 & 0.266 & 33.11 \\
  SwinFusion~\cite{swinfusion} (w. our framework) & 0.813 & 0.639 & \best{0.037} & \best{0.040} & \third{0.269} & \third{35.77} \\ \midruledash
  Ours & \second{0.902} & \second{0.800} & \second{0.058} & \second{0.062} & 0.257 & \best{36.31} \\ 
\bottomrule
\end{tabular}
\end{table}
\vspace{-2mm}
\begin{table}[htbp]
\centering
\caption{Averaged results on lowlight datasets: LOL v1~\cite{lol_v1}, LOL v2~\cite{lol_v2}, and SICE~\cite{sice}, with equal weights between LOL datasets and SICE.
$\dagger$: high no-reference IQA scores coincide with outputs that closely follow the GenAI result, including its hallucinated structures.}
\label{tab:main-lowlight}
\small          
\setlength{\tabcolsep}{3.5pt}  
\begin{tabular}{lcccccc}
\toprule
Experiment & \makecell{\textbf{Content} \\ \textbf{Sim.}$\uparrow$} & \textbf{Qaf}$\uparrow$ & \textbf{W2 Dist.}$\downarrow$ & \makecell{\textbf{W2 Dist.} \\ \textbf{(local)}$\downarrow$} & \textbf{MANIQA}$\uparrow$ & \textbf{MUSIQ}$\uparrow$ \\
\midrule
Input/GenAI output & 0.593 & 0.313 & 0.336 & 0.311 & 0.418 & 66.64 \\ \midruledash
SA-LUT & \second{0.879} & 0.671 & 0.413 & 0.386 & 0.325 & 45.78 \\
Neural Preset & \best{0.988} & \best{0.942} & 0.179 & 0.176 & 0.410 & 55.61 \\
PhotoWCT2 & 0.857 & \second{0.778} & 0.077 & 0.114 & 0.429 & 59.29 \\ \midruledash
CDDFuse (w. fusion losses) & 0.654 & 0.420 & 0.085 & 0.092 & \best{0.488} & \best{66.24} $\dagger$ \\
CDDFuse (w. our framework) & 0.830 & 0.650 & \third{0.069} & \second{0.078} & 0.450 & 63.18 \\
SwinFusion (w. fusion losses) & 0.652 & 0.412 & 0.147 & 0.144 & \second{0.469} & \second{65.94} $\dagger$ \\
SwinFusion (w. our framework) & 0.844 & 0.696 & \best{0.052} & \best{0.060} & 0.430 & 63.31 \\ \midruledash
Ours & \third{0.864} & \third{0.743} & \second{0.065} & \third{0.079} & \third{0.453} & \third{63.34} \\
\bottomrule
\end{tabular}
\end{table}

\vspace{-3mm}
\section{Conclusion}
\label{sec:conclusion}
\vspace{-1mm}
We introduced \emph{structure-preserving GenAI fusion} as a problem setting for black-box GenAI image retouching, and proposed a post-processing framework that addresses four core failure modes: spatial misalignment, texture distortion, content hallucination, and resolution mismatch. Our method combines flow-based spatial alignment, global photometric alignment, and an interpretable multiscale common--unique decomposition network to preserve the structure of the original camera image while inheriting the tonal and perceptual enhancements of the GenAI output. A key property of the framework is its built-in interpretability: the common--unique decomposition exposes a hallucination map as a natural byproduct, which we use to analyze how prompt phrasing affects hallucination rates.

Experiments on tone manipulation and low-light enhancement show that our method achieves a favorable balance between content fidelity, style similarity, and perceptual quality compared to photorealistic style transfer and image fusion baselines. We further demonstrate that our framework is backbone-agnostic: when fusion models are embedded in our pipeline, their content fidelity improves substantially, validating the framework as a general recipe rather than a single-architecture method.

Our framework targets enhancement-based tasks where the camera input is a reliable structural reference. Extending it to settings such as super-resolution or deblurring, where this assumption breaks, remains future work.
\vspace{-2mm}

\bibliographystyle{unsrt}
\bibliography{neurips_2026.bib}

@inproceedings{ma5k,
  author    = {Bychkovsky, Vladimir and Paris, Sylvain and Chan, Eric and Durand, Fr{\'e}do},
  title     = {Learning Photographic Global Tonal Adjustment with a Database of Input/Output Image Pairs},
  booktitle = {Proceedings of the IEEE Conference on Computer Vision and Pattern Recognition (CVPR)},
  year      = {2011},
}

@article{lol_v2,
  author  = {Yang, Wenhan and Wang, Shiqi and Fang, Yuming and Wang, Yue and Liu, Jiaying},
  title   = {Band Representation-Based Semi-Supervised Low-Light Image Enhancement: Bridging the Gap Between Signal Fidelity and Perceptual Quality},
  journal = {IEEE Transactions on Image Processing},
  volume  = {30},
  pages   = {3461--3473},
  year    = {2021},
}

@inproceedings{lol_v1,
  author    = {Wei, Chen and Wang, Wenjing and Yang, Wenhan and Liu, Jiaying},
  title     = {Deep {Retinex} Decomposition for Low-Light Enhancement},
  booktitle = {British Machine Vision Conference (BMVC)},
  year      = {2018},
}

@article{sice,
  author  = {Cai, Jianrui and Gu, Shuhang and Zhang, Lei},
  title   = {Learning a Deep Single Image Contrast Enhancer from Multi-Exposure Images},
  journal = {IEEE Transactions on Image Processing},
  volume  = {27},
  number  = {4},
  pages   = {2049--2062},
  year    = {2018},
}

@inproceedings{csc:lucey,
  author    = {Bristow, Hilton and Eriksson, Anders and Lucey, Simon},
  title     = {Fast Convolutional Sparse Coding},
  booktitle = {Proceedings of the IEEE Conference on Computer Vision and Pattern Recognition (CVPR)},
  year      = {2013},
}

@inproceedings{csc:heide,
  author    = {Heide, Felix and Heidrich, Wolfgang and Wetzstein, Gordon},
  title     = {Fast and Flexible Convolutional Sparse Coding},
  booktitle = {Proceedings of the IEEE Conference on Computer Vision and Pattern Recognition (CVPR)},
  year      = {2015},
}

@inproceedings{csc:deconv,
  author    = {Zeiler, Matthew D. and Krishnan, Dilip and Taylor, Graham W. and Fergus, Rob},
  title     = {Deconvolutional Networks},
  booktitle = {Proceedings of the IEEE Conference on Computer Vision and Pattern Recognition (CVPR)},
  year      = {2010},
}

@inproceedings{lcsc,
  author    = {Sreter, Hillel and Giryes, Raja},
  title     = {Learned Convolutional Sparse Coding},
  booktitle = {Proceedings of the IEEE International Conference on Acoustics, Speech and Signal Processing (ICASSP)},
  year      = {2018},
}

@inproceedings{lista,
  author    = {Gregor, Karol and LeCun, Yann},
  title     = {Learning Fast Approximations of Sparse Coding},
  booktitle = {Proceedings of the International Conference on Machine Learning (ICML)},
  pages     = {399--406},
  year      = {2010},
}

@article{cunet,
  author  = {Deng, Xin and Dragotti, Pier Luigi},
  title   = {Deep Convolutional Neural Network for Multi-Modal Image Restoration and Fusion},
  journal = {IEEE Transactions on Pattern Analysis and Machine Intelligence},
  volume  = {43},
  number  = {10},
  pages   = {3333--3348},
  year    = {2021},
}

@article{lowlevelbanana,
  author  = {Zuo, Jialong and Deng, Haoyou and Zhou, Hanyu and Zhu, Jiaxin and Zhang, Yicheng and Zhang, Yiwei and Yan, Yongxin and Huang, Kaixing and Chen, Weisen and Deng, Yongtai and Jin, Rui and Sang, Nong and Gao, Changxin},
  title   = {Is {Nano Banana Pro} a Low-Level Vision All-Rounder? {A} Comprehensive Evaluation on 14 Tasks and 40 Datasets},
  journal = {arXiv preprint arXiv:2512.15110},
  year    = {2025},
}

@inproceedings{neuralpreset,
  author    = {Ke, Zhanghan and Liu, Yuhao and Zhu, Lei and Zhao, Nanxuan and Lau, Rynson W.~H.},
  title     = {Neural Preset for Color Style Transfer},
  booktitle = {Proceedings of the IEEE/CVF Conference on Computer Vision and Pattern Recognition (CVPR)},
  year      = {2023},
}

@article{scd,
  author  = {Aslantas, Veysel and Bendes, Emre},
  title   = {A New Image Quality Metric for Image Fusion: The Sum of the Correlations of Differences},
  journal = {AEU---International Journal of Electronics and Communications},
  volume  = {69},
  number  = {12},
  pages   = {1890--1896},
  year    = {2015},
}

@inproceedings{w2dist,
  author    = {Lobashev, Alexander and Larchenko, Maria and Guskov, Dmitry},
  title     = {Color Conditional Generation with Sliced {Wasserstein} Guidance},
  booktitle = {Advances in Neural Information Processing Systems (NeurIPS)},
  year      = {2025},
}

@inproceedings{salut,
  author    = {Gong, Zerui and Wu, Zhonghua and Tao, Qingyi and Li, Qinyue and Loy, Chen Change},
  title     = {{SA-LUT}: Spatial Adaptive {4D} Look-Up Table for Photorealistic Style Transfer},
  booktitle = {Proceedings of the IEEE/CVF International Conference on Computer Vision (ICCV)},
  year      = {2025},
}

@inproceedings{photowct2,
  author    = {Chiu, Tai-Yin and Gurari, Danna},
  title     = {{PhotoWCT$^2$}: Compact Autoencoder for Photorealistic Style Transfer Resulting from Blockwise Training and Skip Connections of High-Frequency Residuals},
  booktitle = {Proceedings of the IEEE/CVF Winter Conference on Applications of Computer Vision (WACV)},
  year      = {2022},
}

@inproceedings{wct2,
  author    = {Yoo, Jaejun and Uh, Youngjung and Chun, Sanghyuk and Kang, Byeongkyu and Ha, Jung-Woo},
  title     = {Photorealistic Style Transfer via Wavelet Transforms},
  booktitle = {Proceedings of the IEEE/CVF International Conference on Computer Vision (ICCV)},
  year      = {2019},
}

@inproceedings{photowct,
  author    = {Li, Yijun and Liu, Ming-Yu and Li, Xueting and Yang, Ming-Hsuan and Kautz, Jan},
  title     = {A Closed-Form Solution to Photorealistic Image Stylization},
  booktitle = {Proceedings of the European Conference on Computer Vision (ECCV)},
  year      = {2018},
}

@inproceedings{wct,
  author    = {Li, Yijun and Fang, Chen and Yang, Jimei and Wang, Zhaowen and Lu, Xin and Yang, Ming-Hsuan},
  title     = {Universal Style Transfer via Feature Transforms},
  booktitle = {Advances in Neural Information Processing Systems (NeurIPS)},
  year      = {2017},
}

@inproceedings{deeppreset,
  author    = {Ho, Man M. and Zhou, Jinjia},
  title     = {Deep Preset: Blending and Retouching Photos with Color Style Transfer},
  booktitle = {Proceedings of the IEEE/CVF Winter Conference on Applications of Computer Vision (WACV)},
  year      = {2021},
}

@inproceedings{vgg19,
  author    = {Simonyan, Karen and Zisserman, Andrew},
  title     = {Very Deep Convolutional Networks for Large-Scale Image Recognition},
  booktitle = {International Conference on Learning Representations (ICLR)},
  year      = {2015},
}

@inproceedings{dpst,
  author    = {Luan, Fujun and Paris, Sylvain and Shechtman, Eli and Bala, Kavita},
  title     = {Deep Photo Style Transfer},
  booktitle = {Proceedings of the IEEE Conference on Computer Vision and Pattern Recognition (CVPR)},
  year      = {2017},
}

@inproceedings{flickr2k,
  author    = {Lim, Bee and Son, Sanghyun and Kim, Heewon and Nah, Seungjun and Mu Lee, Kyoung},
  title     = {Enhanced Deep Residual Networks for Single Image Super-Resolution},
  booktitle = {Proceedings of the IEEE Conference on Computer Vision and Pattern Recognition Workshops (CVPRW)},
  year      = {2017},
}

@inproceedings{arbitrary_adain,
  author    = {Huang, Xun and Belongie, Serge},
  title     = {Arbitrary Style Transfer in Real-Time with Adaptive Instance Normalization},
  booktitle = {Proceedings of the IEEE International Conference on Computer Vision (ICCV)},
  year      = {2017},
}

@inproceedings{gatys2016image,
  author    = {Gatys, Leon A. and Ecker, Alexander S. and Bethge, Matthias},
  title     = {Image Style Transfer Using Convolutional Neural Networks},
  booktitle = {Proceedings of the IEEE Conference on Computer Vision and Pattern Recognition (CVPR)},
  year      = {2016},
}

@article{ma2019fusiongan,
  author    = {Ma, Jiayi and Yu, Wei and Liang, Pengwei and Li, Chang and Jiang, Junjun},
  title     = {{FusionGAN}: A Generative Adversarial Network for Infrared and Visible Image Fusion},
  journal   = {Information Fusion},
  volume    = {48},
  pages     = {11--26},
  year      = {2019},
  publisher = {Elsevier},
}

@article{ma2020infrared,
  author    = {Ma, Jiayi and Liang, Pengwei and Yu, Wei and Chen, Chen and Guo, Xiaojie and Wu, Jia and Jiang, Junjun},
  title     = {Infrared and Visible Image Fusion via Detail Preserving Adversarial Learning},
  journal   = {Information Fusion},
  volume    = {54},
  pages     = {85--98},
  year      = {2020},
  publisher = {Elsevier},
}

@article{ganmcc,
  author  = {Ma, Jiayi and Zhang, Hao and Shao, Zhenfeng and Liang, Pengwei and Xu, Han},
  title   = {{GANMcC}: A Generative Adversarial Network with Multiclassification Constraints for Infrared and Visible Image Fusion},
  journal = {IEEE Transactions on Instrumentation and Measurement},
  volume  = {70},
  pages   = {1--14},
  year    = {2021},
}

@article{li2018densefuse,
  author  = {Li, Hui and Wu, Xiao-Jun},
  title   = {{DenseFuse}: A Fusion Approach to Infrared and Visible Images},
  journal = {IEEE Transactions on Image Processing},
  volume  = {28},
  number  = {5},
  pages   = {2614--2623},
  year    = {2018},
}

@article{sdnet,
  author  = {Zhang, Hao and Ma, Jiayi},
  title   = {{SDNet}: A Versatile Squeeze-and-Decomposition Network for Real-Time Image Fusion},
  journal = {International Journal of Computer Vision},
  volume  = {129},
  number  = {10},
  pages   = {2761--2785},
  year    = {2021},
}

@article{DBLP:journals/tip/GaoDXXD22,
  author  = {Gao, Fangyuan and Deng, Xin and Xu, Mai and Xu, Jingyi and Dragotti, Pier Luigi},
  title   = {Multi-Modal Convolutional Dictionary Learning},
  journal = {IEEE Transactions on Image Processing},
  volume  = {31},
  pages   = {1325--1339},
  year    = {2022},
}

@article{DBLP:journals/corr/abs-2005-08448,
  author  = {Xu, Shuang and Zhao, Zixiang and Wang, Yicheng and Zhang, Chunxia and Liu, Junmin and Zhang, Jiangshe},
  title   = {Deep Convolutional Sparse Coding Networks for Image Fusion},
  journal = {arXiv preprint arXiv:2005.08448},
  year    = {2020},
}

@article{deepm2cdl,
  author  = {Deng, Xin and Xu, Jingyi and Gao, Fangyuan and Sun, Xiancheng and Xu, Mai},
  title   = {{DeepM$^2$CDL}: Deep Multi-Scale Multi-Modal Convolutional Dictionary Learning Network},
  journal = {IEEE Transactions on Pattern Analysis and Machine Intelligence},
  volume  = {46},
  number  = {5},
  pages   = {2770--2787},
  year    = {2024},
}

@inproceedings{ddfm,
  author    = {Zhao, Zixiang and Bai, Haowen and Zhu, Yuanzhi and Zhang, Jiangshe and Xu, Shuang and Zhang, Yulun and Zhang, Kai and Meng, Deyu and Timofte, Radu and Van Gool, Luc},
  title     = {{DDFM}: Denoising Diffusion Model for Multi-Modality Image Fusion},
  booktitle = {Proceedings of the IEEE/CVF International Conference on Computer Vision (ICCV)},
  year      = {2023},
}

@inproceedings{cao2024conditional,
  author    = {Cao, Bing and Xu, Xingxin and Zhu, Pengfei and Wang, Qilong and Hu, Qinghua},
  title     = {Conditional Controllable Image Fusion},
  booktitle = {Advances in Neural Information Processing Systems (NeurIPS)},
  year      = {2024},
}

@inproceedings{rffusion,
  author    = {Wang, Zirui and Zhang, Jiayi and Guan, Tianwei and Zhou, Yuhan and Li, Xingyuan and Dong, Minjing and Liu, Jinyuan},
  title     = {Efficient Rectified Flow for Image Fusion},
  booktitle = {Advances in Neural Information Processing Systems (NeurIPS)},
  year      = {2025},
}

@inproceedings{sdxl,
  author    = {Podell, Dustin and English, Zion and Lacey, Kyle and Blattmann, Andreas and Dockhorn, Tim and M{\"u}ller, Jonas and Penna, Joe and Rombach, Robin},
  title     = {{SDXL}: Improving Latent Diffusion Models for High-Resolution Image Synthesis},
  booktitle = {International Conference on Learning Representations (ICLR)},
  year      = {2024},
}

@inproceedings{cddfuse,
  author    = {Zhao, Zixiang and Bai, Haowen and Zhang, Jiangshe and Zhang, Yulun and Xu, Shuang and Lin, Zudi and Timofte, Radu and Van Gool, Luc},
  title     = {{CDDFuse}: Correlation-Driven Dual-Branch Feature Decomposition for Multi-Modality Image Fusion},
  booktitle = {Proceedings of the IEEE/CVF Conference on Computer Vision and Pattern Recognition (CVPR)},
  year      = {2023},
}

@article{swinfusion,
  author  = {Ma, Jiayi and Tang, Linfeng and Fan, Fan and Huang, Jun and Mei, Xiaoguang and Ma, Yong},
  title   = {{SwinFusion}: Cross-Domain Long-Range Learning for General Image Fusion via {Swin} Transformer},
  journal = {IEEE/CAA Journal of Automatica Sinica},
  volume  = {9},
  number  = {7},
  pages   = {1200--1217},
  year    = {2022},
}

@inproceedings{skflow,
  author    = {Sun, Shangkun and Chen, Yuanqi and Zhu, Yu and Guo, Guodong and Li, Ge},
  title     = {{SKFlow}: Learning Optical Flow with Super Kernels},
  booktitle = {Advances in Neural Information Processing Systems (NeurIPS)},
  year      = {2022},
}

@inproceedings{flownet,
  author    = {Dosovitskiy, Alexey and Fischer, Philipp and Ilg, Eddy and H{\"a}usser, Philip and Hazirbas, Caner and Golkov, Vladimir and van der Smagt, Patrick and Cremers, Daniel and Brox, Thomas},
  title     = {{FlowNet}: Learning Optical Flow with Convolutional Networks},
  booktitle = {Proceedings of the IEEE International Conference on Computer Vision (ICCV)},
  pages     = {2758--2766},
  year      = {2015},
}

@inproceedings{maniqa,
  author    = {Yang, Sidi and Wu, Tianhe and Shi, Shuwei and Lao, Shanshan and Gong, Yuan and Cao, Mingdeng and Wang, Jiahao and Yang, Yujiu},
  title     = {{MANIQA}: Multi-Dimension Attention Network for No-Reference Image Quality Assessment},
  booktitle = {Proceedings of the IEEE/CVF Conference on Computer Vision and Pattern Recognition Workshops (CVPRW)},
  year      = {2022},
}

@inproceedings{musiq,
  author    = {Ke, Junjie and Wang, Qifei and Wang, Yilin and Milanfar, Peyman and Yang, Feng},
  title     = {{MUSIQ}: Multi-Scale Image Quality Transformer},
  booktitle = {Proceedings of the IEEE/CVF International Conference on Computer Vision (ICCV)},
  year      = {2021},
}

@inproceedings{nafnet,
  author    = {Chu, Xiaojie and Chen, Liangyu and Yu, Wenqing},
  title     = {{NAFSSR}: Stereo Image Super-Resolution Using {NAFNet}},
  booktitle = {Proceedings of the IEEE/CVF Conference on Computer Vision and Pattern Recognition Workshops (CVPRW)},
  pages     = {1239--1248},
  year      = {2022},
}

@inproceedings{realesrgan,
  author    = {Wang, Xintao and Xie, Liangbin and Dong, Chao and Shan, Ying},
  title     = {{Real-ESRGAN}: Training Real-World Blind Super-Resolution with Pure Synthetic Data},
  booktitle = {Proceedings of the IEEE/CVF International Conference on Computer Vision Workshops (ICCVW)},
  year      = {2021},
}

@article{afifi2025modular,
  author  = {Afifi, Mahmoud and Wang, Zhongling and Zhang, Ran and Brown, Michael S.},
  title   = {Modular Neural Image Signal Processing},
  journal = {arXiv preprint arXiv:2512.08564},
  year    = {2025},
}

@inproceedings{adam,
  author    = {Kingma, Diederik P. and Ba, Jimmy},
  title     = {{Adam}: A Method for Stochastic Optimization},
  booktitle = {International Conference on Learning Representations (ICLR)},
  year      = {2015},
}

@article{instancenorm,
  author  = {Ulyanov, Dmitry and Vedaldi, Andrea and Lempitsky, Victor},
  title   = {Instance Normalization: The Missing Ingredient for Fast Stylization},
  journal = {arXiv preprint arXiv:1607.08022},
  year    = {2016},
}


\clearpage
\appendix

\section{Technical appendices and supplementary material}
\subsection{Method Details}
This section provides additional details on the method described in Sec.~\ref{sec:method}.

\subsubsection{Data synthesis pipeline}
\label{appendix:data_pipeline}
We provide additional details of the data synthesis pipeline used for backbone training.
As described in the main paper, we start from a sampled image patch and apply random color and contrast jittering twice independently to obtain two photometrically different versions of the same scene.
These are used as the input $\mathbf{x}$ and the high-quality target $\mathbf{x}_{hq}$.
Color augmentations include hue and saturation shifts, while brightness and contrast changes are implemented using randomly sampled monotonic 1D tone curves.

The input is then degraded by sequentially applying Gaussian blur, Gaussian noise, and unsharp masking (USM).
To mimic the increased perceptual sharpness often introduced by GenAI retouching, we also apply randomized USM to the target image, with randomly sampled Gaussian blur scale, sharpening strength, and threshold.
We synthesize the GenAI reference $\mathbf{y}$ by applying geometric transformations to the target, including translation, rotation, and shear.
We further simulate content hallucination by copying rectangular regions from the target and pasting them at different locations.
More sophisticated degradation models, such as those used in RealESRGAN~\cite{realesrgan}, could be incorporated in future work for more accurate simulation of real-world camera degradations.

We also apply random rotations, flips, and resizing, where resizing simulates images with different resolutions and fields of view.

\subsubsection{Spatial alignment module}
\label{appendix:spatial_alignment_module}
We use SKFlow~\cite{skflow} as the pretrained optical flow model to spatially align the GenAI output $\mathbf{y}$ with the camera input $\mathbf{x}$.
To reduce computational complexity on full-resolution images, we downsample $\mathbf{x}$ and $\mathbf{y}$ to $512 \times 512$ before performing optical flow.

We perform spatial alignment prior to photometric mapping, as modern flow estimators are inherently robust to the domain gap between $\mathbf{x}$ and $\mathbf{y}$. 
This robustness is largely attributed to aggressive photometric augmentation strategies during training, introducing variations in brightness, contrast, and hue~\cite{flownet, skflow},
and the use of Instance Normalization~\cite{instancenorm}, which removes global brightness biases and contrast variance from the feature space.

Note that the specific choice of optical flow model is not central to our design.
The spatial alignment module may be replaced with any state-of-the-art flow estimator without loss of generality.

\subsubsection{Photometric alignment module}
\label{appendix:photometric_alignment_module}
The photometric alignment module consists of three global operations: tone mapping, a 2D LUT color transform, and gamma correction.

First, global tone mapping is carried out by the following generalized S-curve:
\begin{equation}
\mathbf{x}_{\text{TM}} = g_{\text{TM}}(\mathbf{x}; a, b, c) = \frac{\mathbf{x}^a}{\mathbf{x}^a + (c (1 - \mathbf{x}))^b}, \quad a,b,c=D_{GTM}([\mathbf{x}, \mathbf{y}_{\text{warp}}])
\end{equation}
where $a, b$, and $c$ are parameters controlling the shape of the tone curve, predicted by a CNN-based predictor head $D_{GTM}$.
$[\cdot, \cdot]$ is channel-wise concatenation of the camera image $\mathbf{x}$ with the flow-warped GenAI output $\mathbf{y}_{\text{warp}}$.

The 2D LUT carries out global color transform by mapping the CbCr channels of YCbCr-transformed $\mathbf{x}_{\text{TM}}$ to those of $\mathbf{y}_{\text{warp}}$.
We first separately extract feature maps of the 2D histograms of the CbCr channels of $\mathbf{x}_{\text{TM}}$ and $\mathbf{y}_{\text{warp}}$: 
$\mathbf{h}_{\mathbf{x}_{\text{TM}}} = D_{\text{hist}}(\mathbf{x}_{\text{TM}})$, $\mathbf{h}_{\mathbf{y}_{\text{warp}}} = D_{\text{hist}}(\mathbf{y}_{\text{warp}})$, where $D_{\text{hist}}$ is a shallow convolution-based encoder.
The feature maps are then passed through an encoder-decoder network $D_{\text{chroma}}$ to predict the 2D LUT, as in Eq.~\ref{eq:2dlut}.
The LUT-mapped CbCr values are finally combined with the original Y channel of $\mathbf{x}_{\text{TM}}$ to obtain the color-aligned image $\mathbf{x}_{\text{LUT}}$.
\begin{equation}
\mathbf{x}_{\text{LUT}} = \text{Interp}(\mathbf{x}_{\text{TM}, CbCr}, \text{LUT}_{\text{CbCr}}), \quad \text{LUT}_{\text{CbCr}} = D_{\text{chroma}}([\mathbf{h}_{\mathbf{x}_{\text{TM}}}, \mathbf{h}_{\mathbf{y}_{\text{warp}}}]) 
\label{eq:2dlut} 
\end{equation}

Finally, we gamma-correct $\mathbf{x}_{\text{LUT}}$ with a gamma factor $\gamma$ predicted from $D_{\text{gamma}}$ to further align the brightness level of $\mathbf{x}_{\text{LUT}}$ with $\mathbf{y}_{\text{warp}}$: 
\begin{equation}
\mathbf{x}_{\text{gamma}} = g_{\text{gamma}}(\mathbf{x}_{\text{LUT}}; \gamma) = \mathbf{x}_{\text{LUT}}^\gamma, \quad \gamma = D_{\text{gamma}}([\mathbf{x}_{\text{LUT}}, \mathbf{y}_{\text{warp}}])  
\end{equation}

For more details on each sub-network, please refer to~\cite{afifi2025modular}.

\textbf{Data.}
Our backbone data synthesis pipeline encourages robustness to fine-grained textural distortions, 
but does not explicitly cover larger hallucinated content, such as hallucinated clouds in the sky.
To train the photometric alignment module, we therefore augment the main dataset with 50 Nano Banana Pro-generated examples containing such hallucinations.
This helps the module learn color and brightness matching while disregarding large hallucinated regions in the GenAI output.

\textbf{Training.}
We train the photometric alignment module with a frozen flow estimator applied beforehand, 
so that the module takes the warped GenAI output $\mathbf{y}_{\text{warp}}$ as input.
This prevents spatial misalignment from confounding the prediction of color and tone mappings.

\subsubsection{Fusion module}
\label{appendix:fusion_module}
\begin{figure}[t]
    \centering
    \includegraphics[width=1.0\linewidth]{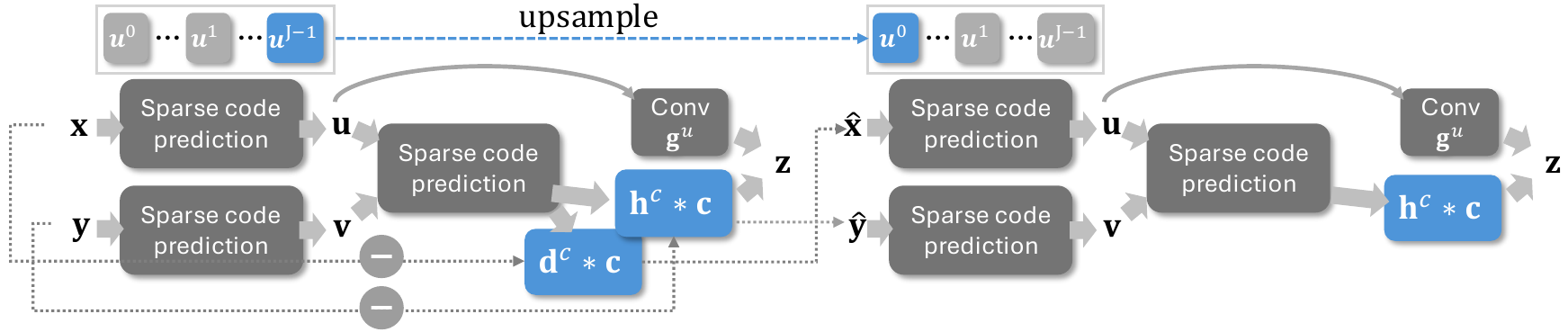}
    \caption{Model architecture. We show only two scales for simplicity. 
    The coarsest scale takes in the original input $\mathbf{x}$ and the GenAI reference $\mathbf{y}$, 
    and predicts the common and unique codes $\mathbf{c}, \mathbf{u}, \mathbf{v}$. 
    The output is reconstructed from the image-space common component of $\mathbf{y}$ and the unique code $\mathbf{u}$ of the input $\mathbf{x}$. 
    The finer scale takes the upsampled sparse codes from the coarser scale as warm-start, 
    and also takes the unique-component residuals $\mathbf{\hat{x}}, \mathbf{\hat{y}}$ as inputs to predict finer-scale sparse codes.}
    \label{fig:model}
\end{figure}
We show a diagram of the proposed multiscale common-unique decomposition network (MSCU-Net) in Figure~\ref{fig:model}.
For reference, we summarize the symbols used in the network and their corresponding interpretations in Table~\ref{tab:notation},
and summarize the equations in Eq.~\ref{eq:mscunet}.
In this section, $\mathbf{x}$ and $\mathbf{y}$ denote two generic source images.
Within our framework, they correspond to $\mathbf{x}_{\text{photo}}$ and $\mathbf{y}_{\text{warp}}$, respectively.
\begin{table}[t]
\centering
\caption{Notation summary for the common-unique decomposition framework.}
\small
\setlength{\tabcolsep}{6pt}
\begin{tabular}{ll}
\toprule
\textbf{Symbol} & \textbf{Description} \\
\midrule
$\mathbf{u}^{(j)}$ & Unique sparse code for $\mathbf{x}$ at LISTA~\cite{lcsc, lista} iteration $j$ \\
$\mathbf{v}^{(j)}$ & Unique sparse code for $\mathbf{y}$ at LISTA iteration $j$ \\
$\mathbf{c}^{(j)}$ & Common sparse code at LISTA iteration $j$ \\
$\mathbf{d}^c$ & Dictionary associated with common sparse code \\
$\mathbf{d}^u_e,\ \mathbf{d}^u_d\ (\equiv \mathbf{d}^u)$  & Encoding and decoding dictionaries for unique sparse code of $\mathbf{x}$ \\
$\mathbf{h}^v_e,\ \mathbf{h}^v_d\ (\equiv \mathbf{h}^v)$ & Encoding and decoding dictionaries for unique sparse code of $\mathbf{y}$ \\
$\mathbf{w}^c_{d} \equiv[\mathbf{d}^{c}, \mathbf{h}^{c}]$ & Concatenated decoding dictionaries for common sparse code $\mathbf{c}$ \\
$\mathbf{d}^c * \mathbf{c},\ \mathbf{h}^c * \mathbf{c}$ & Image-space common components of $\mathbf{x}$ and $\mathbf{y}$, respectively \\
$\mathbf{d}^u * \mathbf{u},\ \mathbf{h}^v * \mathbf{v}$ & Image-space unique components of $\mathbf{x}$ and $\mathbf{y}$, respectively \\
$\mathbf{\hat{x}}=\mathbf{x} - \mathbf{d}^c * \mathbf{c}$ & Image-space unique residual for $\mathbf{x}$ \\
$\mathbf{\hat{y}}=\mathbf{y} - \mathbf{h}^c * \mathbf{c}$ & Image-space unique residual for $\mathbf{y}$ \\
$\mathbf{\tilde{x}}=\mathbf{x} - \mathbf{d}^u * \mathbf{u}$ & Image-space common residual associated with $\mathbf{x}$ \\
$\mathbf{\tilde{y}}=\mathbf{y} - \mathbf{h}^v * \mathbf{v}$ & Image-space common residual associated with $\mathbf{y}$ \\
$\mathbf{z}$ & Fused output \\
$\mathbf{g}^u$ & Dictionary for $\mathbf{z}$ associated with unique sparse code of $\mathbf{x}$ \\
\bottomrule
\end{tabular}

\label{tab:notation}
\end{table}

\begin{equation}
\begin{aligned}
&\text{For each scale } s \text{ (omitted for readability)}: \\
&\mathbf{u}^{(j+1)} = S_{\theta_j^u} \left( \mathbf{u}^{(j)} - \mathbf{d}_e^u * \mathbf{d}_d^u * \mathbf{u}^{(j)} + \mathbf{d}_e^u * \mathbf{\hat{x}} \right) \\
&\mathbf{v}^{(j+1)} = S_{\theta_j^v}\left( \mathbf{v}^{(j)} - \mathbf{h}_e^v * \mathbf{h}_d^v * \mathbf{v}^{(j)} + \mathbf{h}_e^v * \mathbf{\hat{y}} \right) \\
&\mathbf{c}^{(j+1)} = S_{\theta_j^c}\left( \mathbf{c}^{(j)} - \mathbf{w}_e^c * \mathbf{w}_d^c * \mathbf{c}^{(j)} + \mathbf{w}_e^c * [\mathbf{\tilde{x}}, \mathbf{\tilde{y}}] \right)  \\
&\mathbf{u}^{(0),(s+1)}= \uparrow \mathbf{u}^{(J-1),(s)} \\
&\mathbf{v}^{(0),(s+1)}= \uparrow \mathbf{v}^{(J-1),(s)} \\
&\mathbf{c}^{(0),(s+1)}= \uparrow \mathbf{c}^{(J-1),(s)} \\
&\mathbf{z}^{(s)}= \mathbf{h}^{c,(s)} * \mathbf{c}^{(J-1),(s)} + \mathbf{g}^{u,(s)} * \mathbf{u}^{(J-1),(s)}, \\ 
&\mathbf{\hat{x}}^{(0)}=\mathbf{x}^{(0)} \\
&\mathbf{\hat{y}}^{(0)}=\mathbf{y}^{(0)} \\
&\hat{\mathbf{x}}^{(s+1)}=\mathbf{x}^{(s+1)} - \uparrow (\mathbf{d}^{c,(s)} * \mathbf{c}^{(s)}) \\
&\mathbf{\hat{y}}^{(s+1)}=\mathbf{y}^{(s+1)} - \uparrow (\mathbf{h}^{c,(s)} * \mathbf{c}^{(s)}) \\
&\mathbf{\tilde{x}}^{(s)}=\mathbf{x}^{(s)} - \mathbf{d}^{u,(s)} * \mathbf{u}^{(J-1),(s)} \\
&\mathbf{\tilde{y}}^{(s)}=\mathbf{y}^{(s)} - \mathbf{h}^{v,(s)} * \mathbf{v}^{(J-1),(s)}
\end{aligned}    
\label{eq:mscunet} 
\end{equation}

Aside from the multiscale architecture, 
we also modify the output reconstruction to directly use the common reconstruction for $\mathbf{y}$, $\mathbf{h}^{c} * \mathbf{c}$:
\begin{equation}
\mathbf{z}^{(s)}=\mathbf{h}^{c,(s)} * \mathbf{c}^{(s)} + \mathbf{g}^{u,(s)} *\mathbf{u}^{(s)} \label{eq:fuse}   
\end{equation}
This design bases the reconstruction on the GenAI reference, allowing the output to directly inherit its colors and tones.
Since $\mathbf{h}^{c,(s)} * \mathbf{c}^{(s)}$ is optimized to reconstruct the common content of $\mathbf{y}$, it is expected to capture the desired photometric characteristics of the GenAI reference.
Meanwhile, we exclude the GenAI-unique component $\mathbf{v}^{(s)}$ during fusion to suppress hallucinated content.

\textbf{Hallucination map.}
The explicit common-unique decomposition enables a simple hallucination map derived from the GenAI-unique component.
We obtain this component in image space as
$\mathbf{h}_m = \mathbf{h}^{v,(s)} * \mathbf{v}^{(s)}$.
Since hallucinated content in $\mathbf{y}$ is unsupported by the shared scene structure, 
it is expected to be weakly represented by the common code $\mathbf{c}$ and instead captured by the unique code $\mathbf{v}$.
We thus use the activation magnitude of $\mathbf{h}_m$ as a proxy for hallucinated content, 
computing the map as the per-pixel $\ell_2$ norm normalized to $[0,1]$:
\begin{equation}
  \mathbf{m} = \|\mathbf{h}_m\|_2, \quad
  \hat{\mathbf{m}} = \mathbf{m} / \max(\mathbf{m}).
\end{equation}

\subsubsection{Residual refinement}
\label{appendix:nafnet_addon}
The multiscale CU-Net provides an interpretable common-unique decomposition that supports hallucination-aware fusion.
However, the constraints imposed by this decomposition can limit fine-detail reconstruction compared with generic restoration architectures.
To improve visual quality while preserving the decomposition-based prediction, we add a lightweight NAFNet~\cite{nafnet} refinement module after the CU-Net output.
The refiner is intended to improve residual low-level details rather than replace the interpretable fusion stage.

The refiner takes the aligned input image $\mathbf{x}_{\text{photo}}$, warped GenAI reference $\mathbf{y}_{\text{warp}}$, 
preliminary fused output from MSCU-Net $\mathbf{z}^{(2)}$ (output at the finest scale), and the hallucination map $\mathbf{\hat{m}}$ as input.
The preliminary output and hallucination map are detached, so the refiner acts as a residual correction stage without altering the learned decomposition.
Formally, the final output is
\begin{equation}
    \mathbf{z}
    =
    \mathbf{y}_{\text{warp}}
    +
    R_{\theta}
    \left([
    \mathbf{x}_{\text{photo}},
    \mathbf{y}_{\text{warp}},
    \mathrm{sg}(\mathbf{z}^{(2)}),
    \mathrm{sg}(\hat{\mathbf{m}})]
    \right),
\end{equation}
where $\mathrm{sg}(\cdot)$ denotes stop-gradient.

\subsection{Experiments}
\label{appendix:experiments}

\subsubsection{Implementation details}
This section provides more details on hyperparameters.

\textbf{Data synthesis pipeline hyperparameters.}
For input degradation, we apply Gaussian blur, Gaussian noise, and USM sharpening, each with probability $0.5$.
Gaussian blur uses $\sigma \in [0.5,1.5]$; 
Gaussian noise uses $\sigma_n \in [0.002,0.05]$; 
and USM uses kernel sizes in $[3,15]$ with strength in $[1,3]$.
For detail augmentation for the target, USM is applied with probability $0.5$ using kernel sizes in $[3,15]$ and strength in $[1,5]$.
Reference (simulated GenAI output) hallucinations are simulated with homography perturbations and cut-paste augmentation: 
translation is bounded by $5\%$ of the image size, rotation by $5^\circ$, shear by $0.05$, scale by $[0.95,1.05]$, 
and copy-paste patches occupy $5\%$--$40\%$ of the image size.

\textbf{Training hyperparameters.}
As described in the main paper, we first train MSCU-Net and the photometric alignment module (PAM) separately.
MSCU-Net is trained using the data synthesis pipeline in Sec.~\ref{appendix:data_pipeline}, with $256 \times 256$ patches and a batch size of 8.
We set the weighting factor $\alpha$ to $0.5$ for the loss at coarser scales.
PAM is trained following Sec.~\ref{appendix:photometric_alignment_module}, using patches with sizes uniformly sampled from $\{256,512\}$.
The larger patch size is included because PAM predicts global photometric mappings, which require broader image context than local texture modeling.

After pretraining, we finetune MSCU-Net together with PAM for 50 epochs using the Adam optimizer~\cite{adam} with learning rate 1e-5.
This stage uses only the 50-image large-hallucination dataset, allowing MSCU-Net to better distinguish large hallucinated regions from valid GenAI enhancements.
With PAM integrated, this distinction becomes easier: after photometric alignment, 
unsupported hallucinated content often appears as localized color or brightness discrepancies between the mapped input and the GenAI reference.

For residual refinement, we use a compact NAFNet with width 10, one middle block, and $[1,1,1,1]$ encoder and decoder blocks.
We train the cascaded model for 200 epochs with learning rate $0.001$.

\textbf{Training and evaluation datasets.}
For testing on the MA5K dataset~\cite{ma5k}, we use 285 manually-selected test scenes from the 500-image test split provided by~\cite{lowlevelbanana}.
We randomly divide the remaining 4500 images into 4000 training and 500 validation images.
For testing on the low-light datasets, we directly use the input--NB Pro pairs provided by~\cite{lowlevelbanana}.

\subsubsection{Framework ablations}
\label{appendix:ablations}

\begin{figure}[t]
    \centering
    \includegraphics[width=0.95\linewidth]{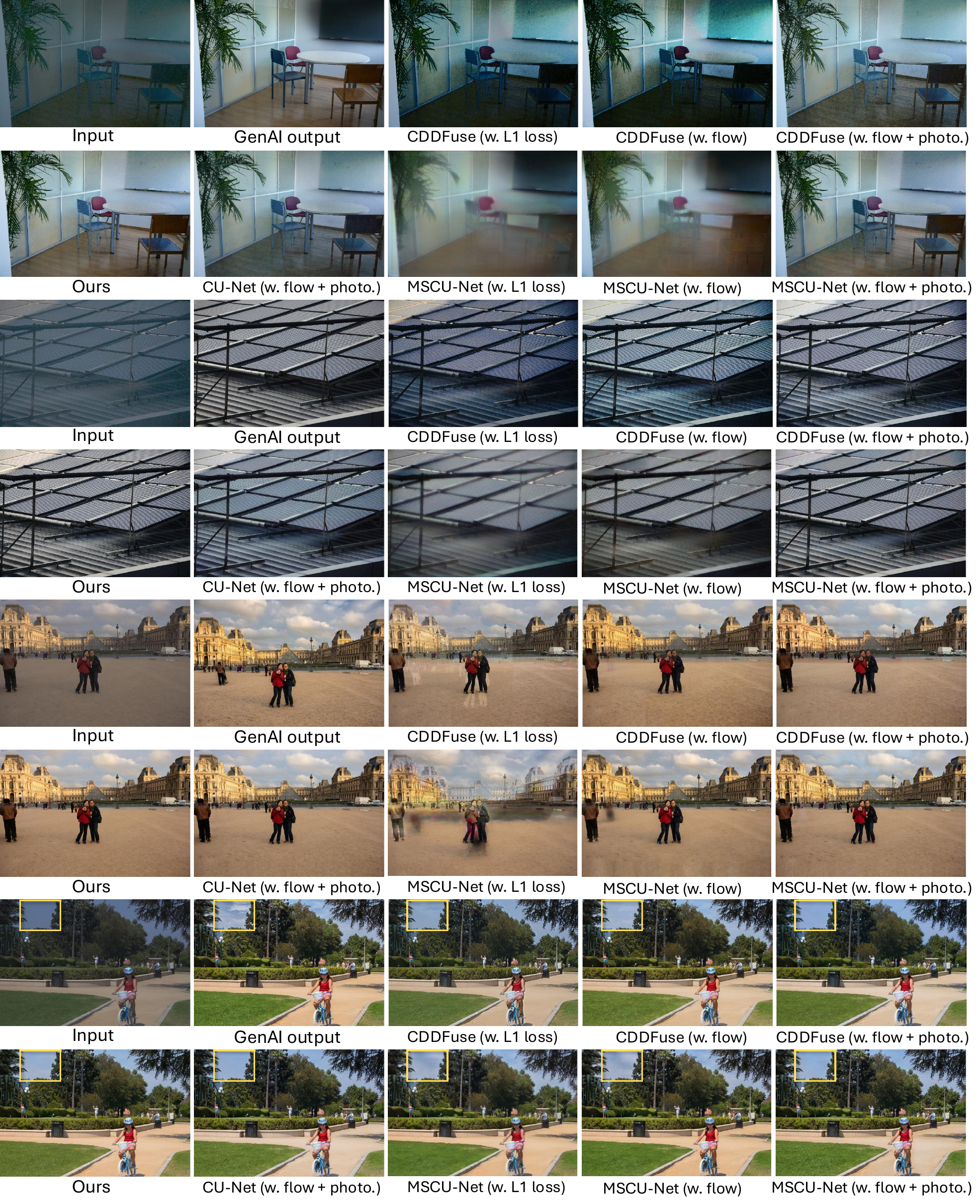}
    \caption{Qualitative results of ablations on MIT-Adobe FiveK~\cite{ma5k} and LOL v2~\cite{lol_v2}.}
    \label{fig:ablations}
\end{figure}

Ablation results are shown in Table~\ref{tab:ablation-lowlight} and Figure~\ref{fig:ablations}.
We evaluate the effect of our framework on three fusion backbones: CDDFuse~\cite{cddfuse}, SwinFusion~\cite{swinfusion}, and the multiscale common-unique decomposition network (MSCU-Net).

Backbones trained with only a reconstruction loss often show low style similarity to the GenAI reference and low perceptual quality, likely due to artifacts caused by spatial misalignment (see Figure~\ref{fig:ablations}, row 5, column 3).
Adding a frozen flow estimator reduces these artifacts, but the outputs may still fail to match the tonal characteristics of the GenAI reference (see rows 1 and 3, column 4).
Moreover, flow alone does not fully address large-scale hallucinations (see rows 7 and 8, column 4).

With both flow and the photometric alignment module, the models more faithfully match the GenAI reference in tone while better suppressing hallucinated content.

\begin{table}[htbp]
\centering
\caption{Ablation results on LOL v2~\cite{lol_v2}.
The \besttext{best}, \secondtext{second}, and \thirdtext{third} results are highlighted.
(w. L1 loss): backbone model trained with L1 loss only.
(w. flow): backbone model cascaded with a pretrained flow estimator.
(w. flow + photo.): backbone model cascaded with a pretrained flow estimator and the photometric alignment module.
}
\label{tab:ablation-lowlight}
\small          
\setlength{\tabcolsep}{3.5pt}
\begin{tabular}{lcccccc}
\toprule
Experiment & \makecell{\textbf{Content} \\ \textbf{Sim.}$\uparrow$} & \textbf{Qaf}$\uparrow$ & \textbf{W2 Dist.}$\downarrow$ & \makecell{\textbf{W2 Dist.} \\ \textbf{(local)}$\downarrow$} & \textbf{MANIQA}$\uparrow$ & \textbf{MUSIQ}$\uparrow$ \\
\midrule
CDDFuse~\cite{cddfuse} (w. L1 loss) & \best{0.894} & \third{0.710} & \third{0.136} & \third{0.279} & \third{0.398} & \third{52.02} \\
CDDFuse~\cite{cddfuse} (w. flow) & \second{0.892} & \best{0.734} & \second{0.093} & \second{0.306} & \second{0.431} & \second{55.37} \\
CDDFuse~\cite{cddfuse} (w. flow + photo.) & \third{0.887} & \second{0.728} & \best{0.066} & \best{0.317} & \best{0.446} & \best{56.73} \\ \midruledash

SwinFusion~\cite{swinfusion} (w. L1 loss) & \third{0.797} & \second{0.605} & \second{0.114} & \third{0.272} & \third{0.366} & \third{53.36} \\
SwinFusion~\cite{swinfusion} (w. flow) & \second{0.813} & \third{0.596} & \third{0.136} & \best{0.320} & \second{0.421} & \second{56.02} \\
SwinFusion~\cite{swinfusion} (w. flow + photo.) & \best{0.844} & \best{0.651} & \best{0.051} & \second{0.312} & \best{0.446} & \best{58.22} \\ \midruledash

MSCU-Net (w. L1 loss) & \third{0.751} & \third{0.412} & \third{0.069} & \third{0.080} & \third{0.257} & \third{42.78} \\
MSCU-Net (w. flow) & \second{0.777} & \second{0.474} & \second{0.063} & \second{0.065} & \second{0.451} & \second{59.77} \\
MSCU-Net (w. flow + photo.) & \best{0.817} & \best{0.564} & \best{0.054} & \best{0.055} & \best{0.461} & \best{61.05} \\ 
\bottomrule
\end{tabular}
\end{table}

\begin{figure}[t]
    \centering
    \includegraphics[width=1.0\linewidth]{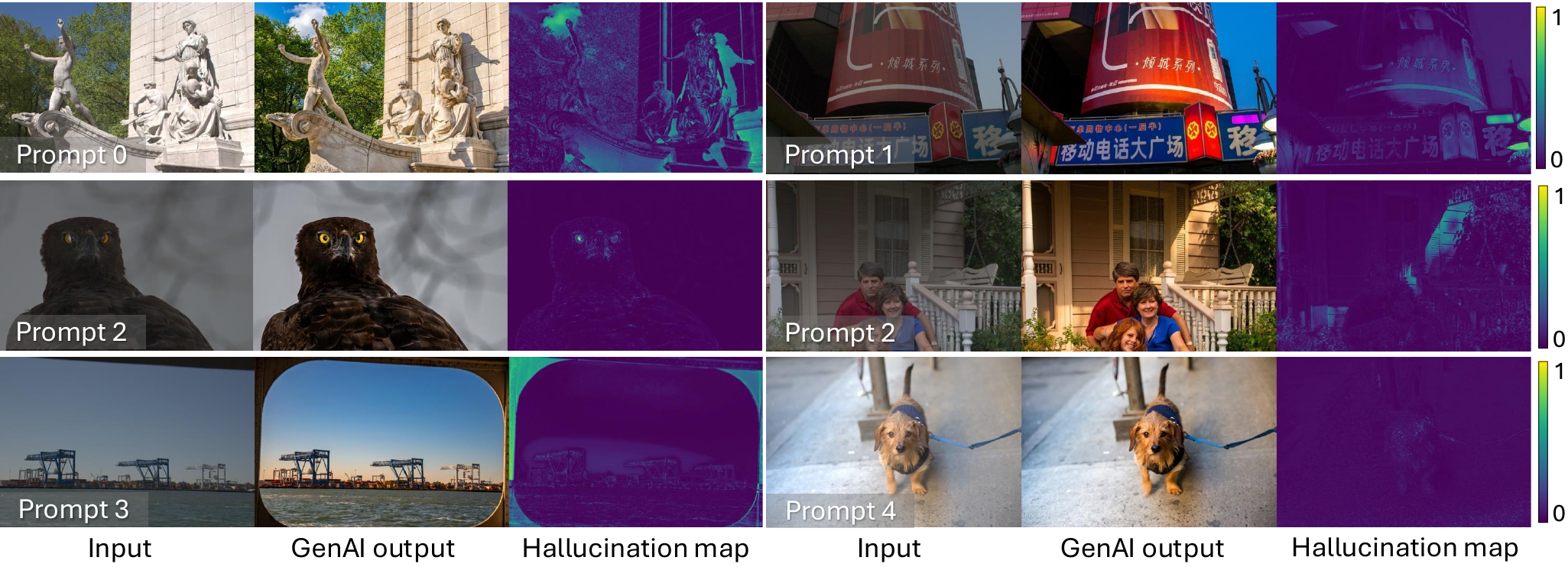}
    \caption{
    Representative examples from the 40-image dataset used for prompt sensitivity analysis.
    Activated regions in the hallucination map indicate content mismatches between the input and GenAI output.
    When the GenAI output remains structurally consistent with the input, the hallucination map shows few activations, as shown in row 3, last column.
    }
    \label{fig:halluc_map_for_prompt}
\end{figure}

\subsubsection{Fusion model interpretability}
\label{appendix:interpretability}

\begin{figure}[t]
    \centering
    \includegraphics[width=1.0\linewidth]{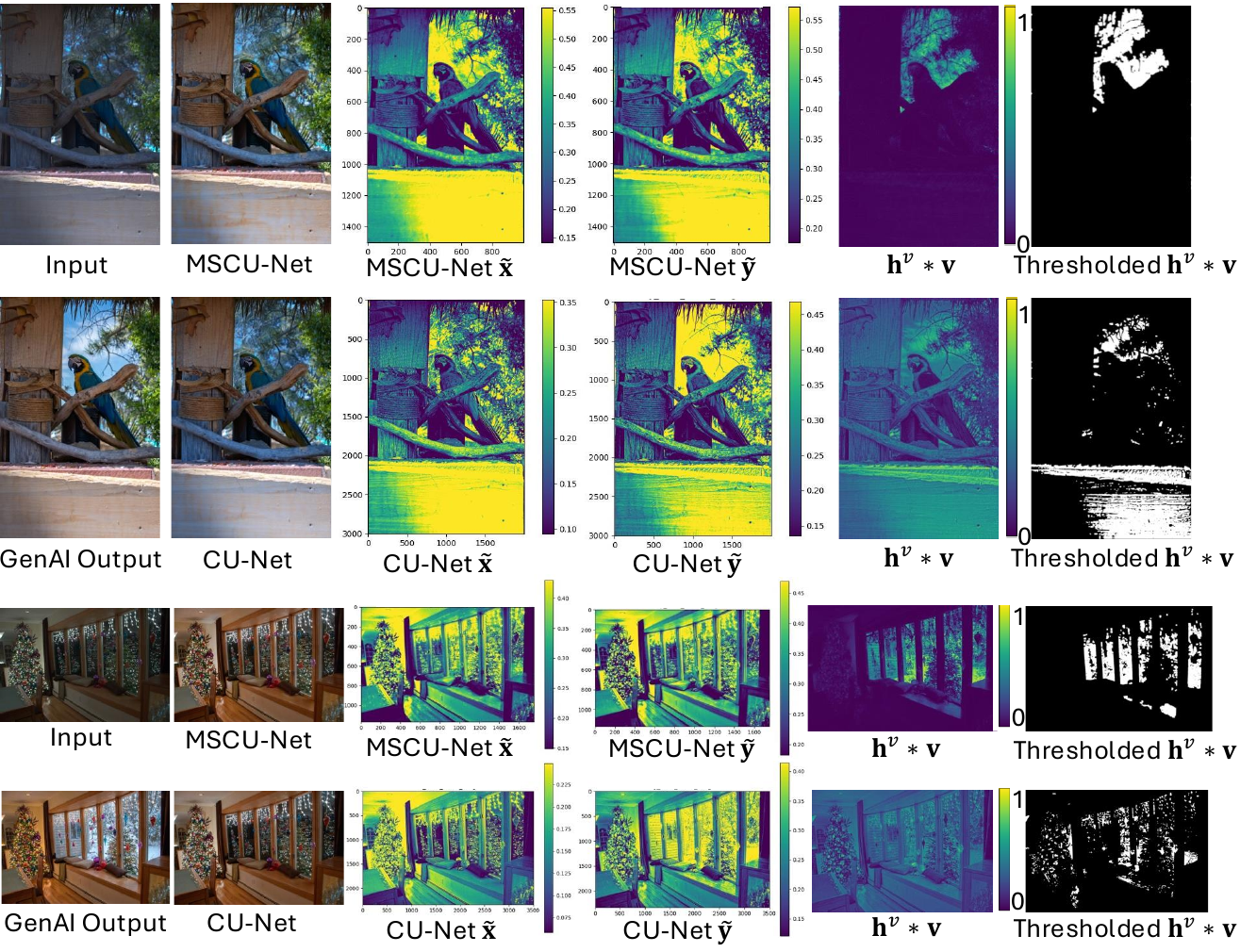}
    \caption{Visualizations of CU-Net's~\cite{cunet} and MSCU-Net's intermediate feature maps and hallucination maps.
    From left to right, we show the input/GenAI image, fused result, common residuals, unique component of GenAI image as hallucination map, and thresholded unique component.}
    \label{fig:halluc_map_cunet_vs_ours}
\end{figure}

Figure~\ref{fig:halluc_map_cunet_vs_ours} compares the intermediate decompositions produced by CU-Net~\cite{cunet} and our MSCU-Net.
The middle columns visualize $\tilde{\mathbf{x}}$ and $\tilde{\mathbf{y}}$, the image-space common residuals of the input and GenAI reference, respectively, which are used as inputs to the common-code predictor in Eq.~\ref{eq:ista_c}.
We visualize the normalized $\ell_2$ magnitude maps.
For the CU-Net baseline, $\tilde{\mathbf{x}}$ and $\tilde{\mathbf{y}}$ tend to retain structures specific to their corresponding source images, rather than emphasizing a shared component.
In contrast, our MSCU-Net produces more consistent common components across the two inputs, suggesting a cleaner common-unique decomposition and improved interpretability.

The right columns visualize the GenAI-unique image-space component, $\mathbf{h}_m = \mathbf{h}^{v} * \mathbf{v}$.
We show both the normalized magnitude map $\hat{\mathbf{m}}$ and a thresholded version.
Compared with CU-Net, our MSCU-Net produces a more localized response on regions that are unsupported by the input, yielding a clearer hallucination map.

\subsubsection{Additional results}

\begin{figure}[t]
    \centering
    \includegraphics[width=1.0\linewidth]{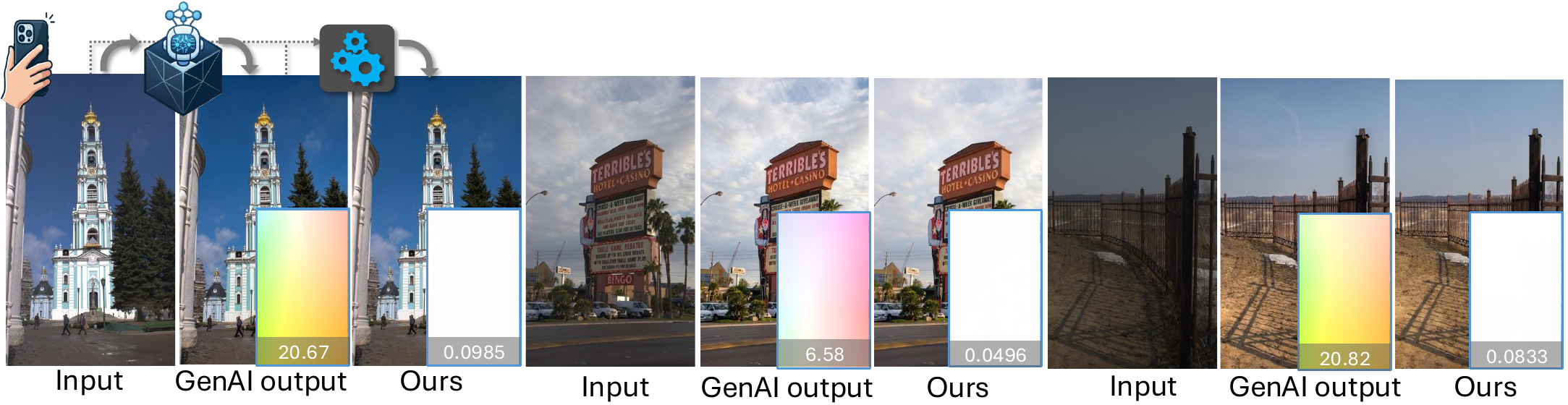}
    \caption{
        Optical-flow visualization relative to the input.
        The GenAI output shows noticeable spatial misalignment, whereas our post-processed result is nearly aligned with the input, producing near-zero flow magnitude.
    }
    \label{fig:results_flow}
\end{figure}

\begin{figure}[t]
    \centering
    \includegraphics[width=0.95\linewidth]{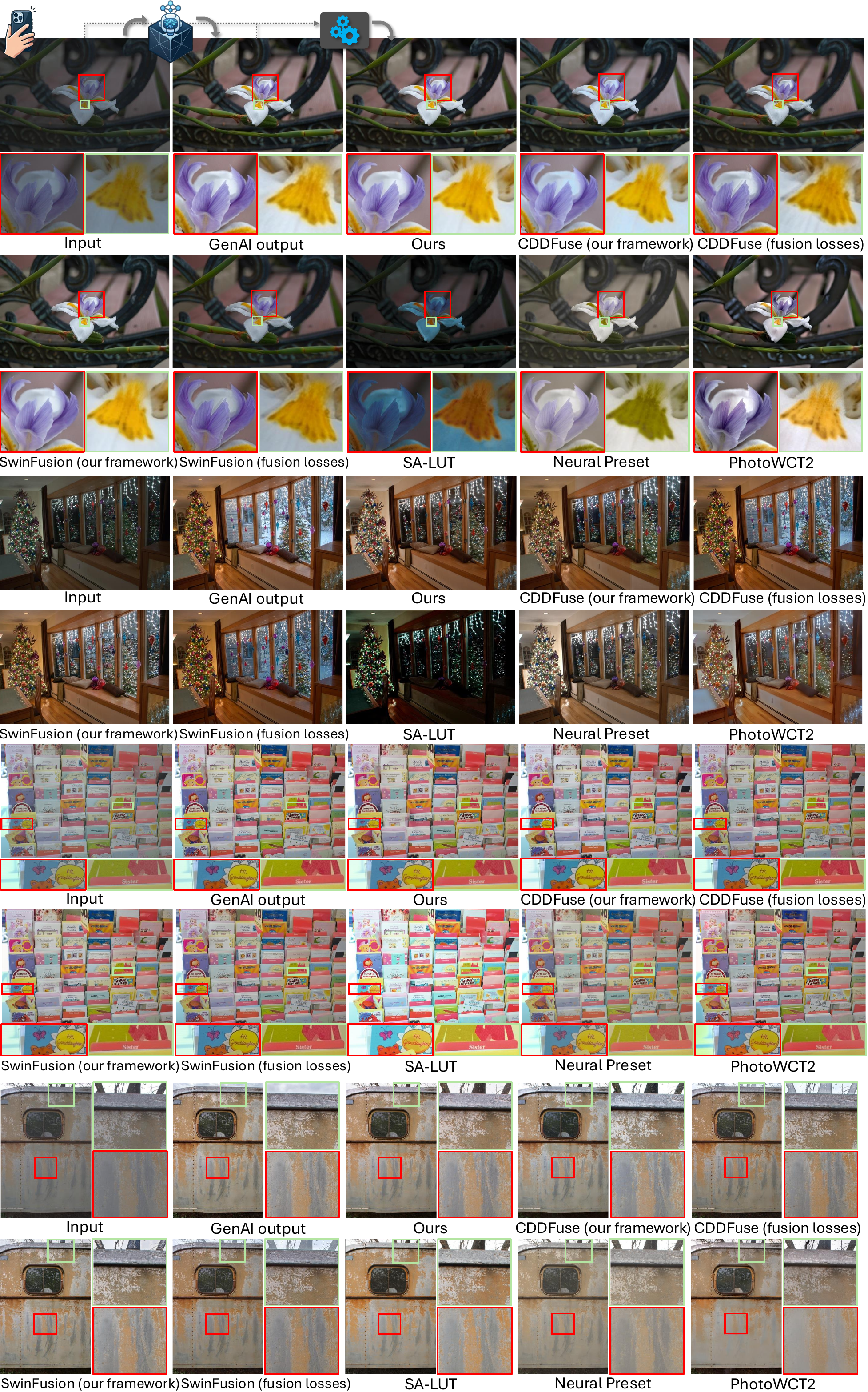}
    \caption{Qualitative results on NB Pro MIT-Adobe FiveK~\cite{ma5k}.}
    \label{fig:qualitative_ma5k_2_3}
\end{figure}

\begin{figure}[t]
    \centering
    \includegraphics[width=1.0\linewidth]{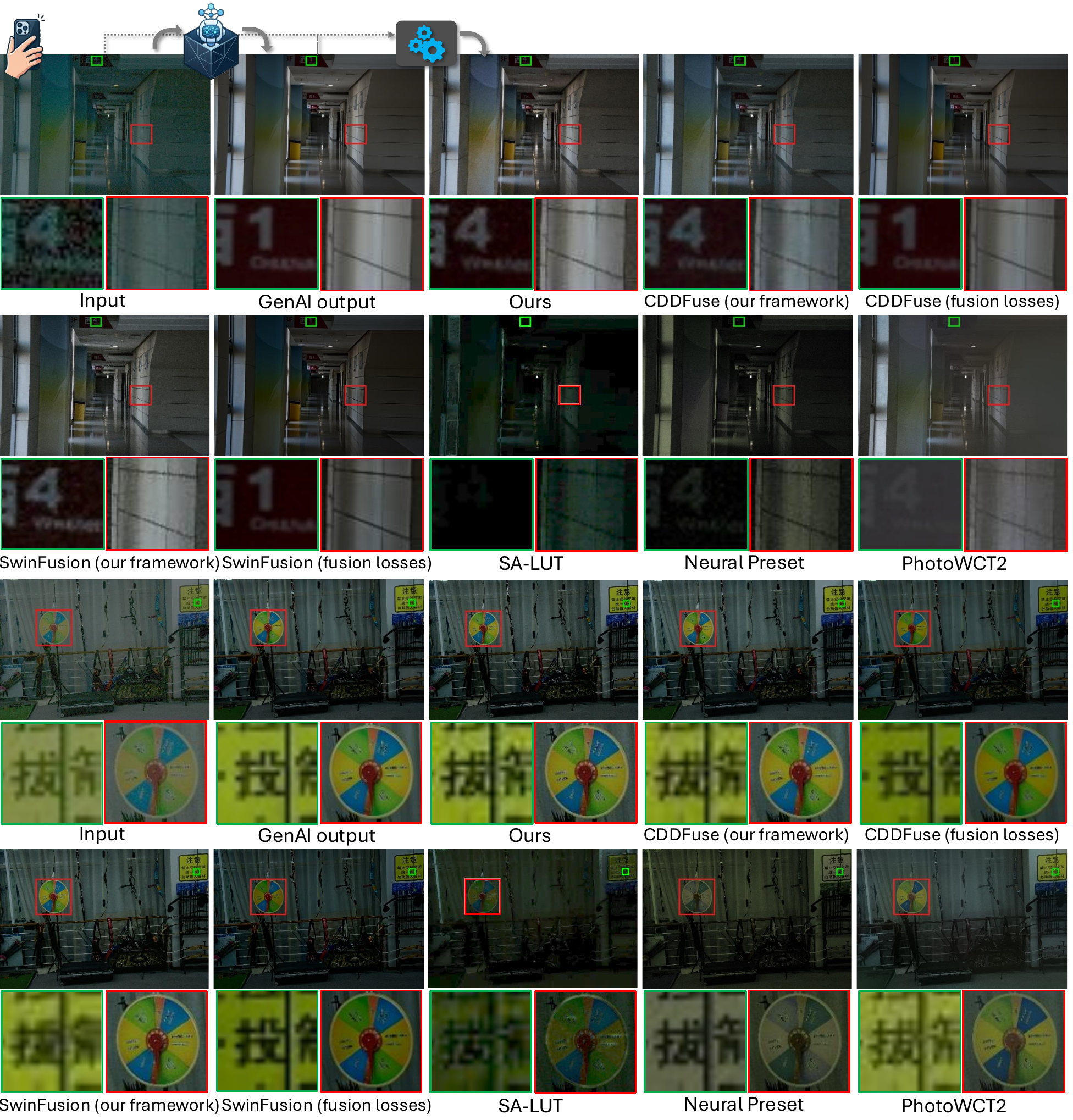}
    \caption{Qualitative results on NB Pro LOL v2~\cite{lol_v2}.}
    \label{fig:qualitative_lol_v2_2}
\end{figure}

\textbf{Additional qualitative results.} We show additional qualitative results on the Nano Banana Pro evaluation datasets in Figures~\ref{fig:qualitative_ma5k_2_3},~\ref{fig:qualitative_lol_v2_2}, 
and~\ref{fig:results_flow}.

\textbf{Evaluation on ChatGPT-generated data.}
To test whether our method generalizes beyond the Nano Banana Pro data, 
we further evaluate it on outputs from another black-box GenAI model, ``gpt-image-1.5'', without retraining.
Specifically, we use ``gpt-image-1.5'' to enhance 493 scenes from the 500-image MIT-Adobe FiveK~\cite{ma5k} test split used in~\cite{lowlevelbanana} with the tone manipulation prompt, 
and 100 LOL v2~\cite{lol_v2} scenes with the low-light enhancement prompt.
Results are reported in Table~\ref{tab:chatgpt}.

\begin{figure}[b]
    \centering
    \includegraphics[width=1.0\linewidth]{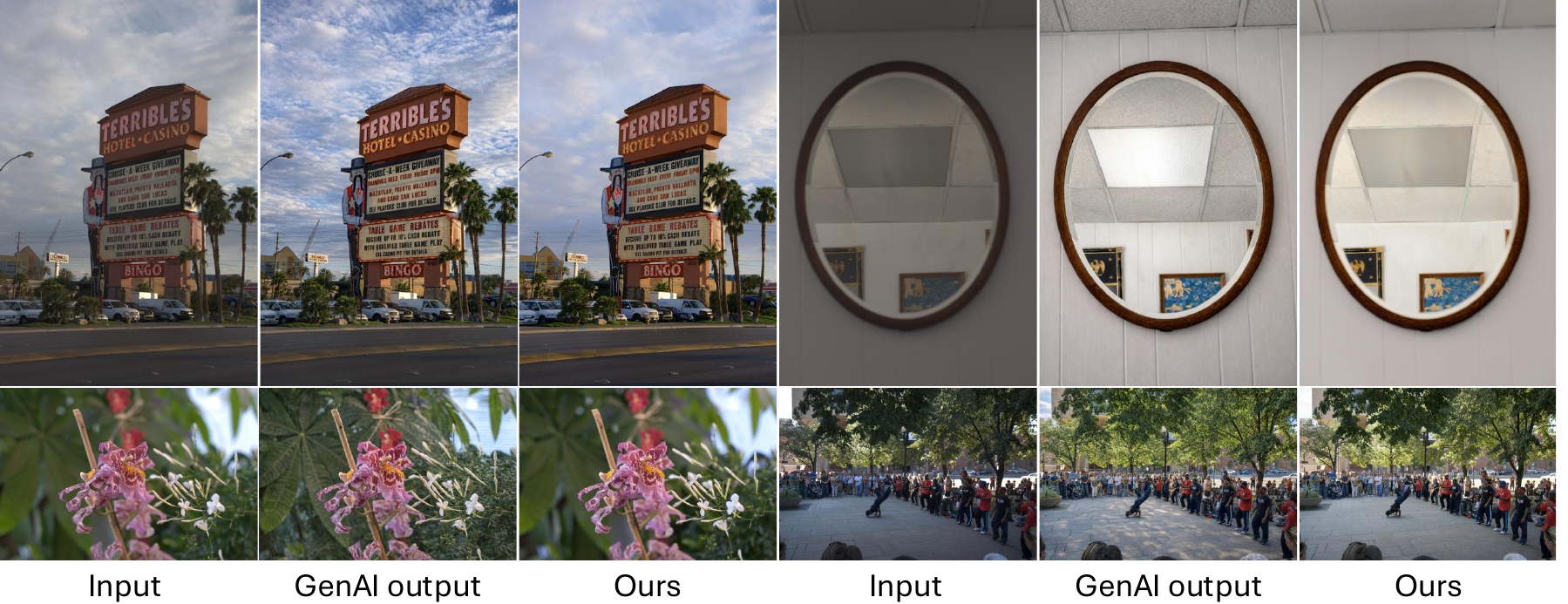}
    \caption{Qualitative results on ChatGPT MIT-Adobe FiveK~\cite{ma5k}.}
    \label{fig:qualitative_chatgpt}
\end{figure}

For content fidelity, we compare the GenAI output and our result against the input image.
For style similarity, we compare the input and our result against the GenAI output.
No-reference IQA metrics are reported for the input, GenAI output, and our result.
Compared with the GenAI output, our method substantially improves content fidelity while retaining the desired enhancement effects.
Qualitative examples are shown in Figure~\ref{fig:qualitative_chatgpt}.

\begin{table}[htbp]
\centering
\caption{Results on MIT-Adobe FiveK~\cite{ma5k} and LOL v2~\cite{lol_v2} using ChatGPT ``gpt-image-1.5'' as the GenAI model.}
\label{tab:chatgpt}
\small
\setlength{\tabcolsep}{3.5pt}
\begin{tabular}{lcccccc}
\toprule
& \makecell{\textbf{Content} \\ \textbf{Sim.}$\uparrow$} & \textbf{Qaf}$\uparrow$ & \textbf{W2 Dist.}$\downarrow$ & \textbf{MANIQA}$\uparrow$ & \textbf{MUSIQ}$\uparrow$ \\
\midrule
\multicolumn{6}{l}{\textit{MIT-Adobe FiveK}} \\
\midrule
Input          & --    & --    & 0.209    & 0.260 & 35.32 \\
GenAI output   & 0.461 & 0.177 & --       & 0.288 & 35.35 \\
Ours           & 0.906 & 0.788 & 0.066    & 0.250 & 35.91 \\ \midruledash
\multicolumn{6}{l}{\textit{LOL v2}} \\
\midrule
Input           & --     & --    & 0.737    & 0.321 & 39.14 \\
GenAI output    & 0.525  & 0.277 & --       & 0.498 & 71.33 \\
Ours            & 0.790  & 0.681 & 0.062    & 0.461 & 60.59 \\
\bottomrule
\end{tabular}
\end{table}

\subsubsection{Model complexity and inference cost}
\label{appendix:complexity}
Table~\ref{tab:model_specs} reports the model complexity and inference cost of the baselines and our method.
All models are trained and evaluated using a single NVIDIA L40 GPU with 48 GB of memory.
We observe that image fusion baselines can have a high memory footprint despite their modest parameter counts, 
as they generate multiple full-resolution feature maps during inference.
This makes direct inference on large camera images difficult and may require tiled inference, which can introduce visible artifacts.
Our method maintains a comparatively lower memory footprint while achieving competitive runtime, making it more suitable for full-resolution inputs.

\begin{table}[htbp]
\centering
\caption{
Model complexity and inference cost. 
Inference time and MACs are measured at $1024 \times 1024$.
Memory is measured at $1024 \times 1024$ and full-resolution $3000 \times 4000$ inputs.
``Failed'' indicates that direct full-resolution inference did not complete, due to either out-of-memory or runtime errors.
}
\label{tab:model_specs}
\small
\setlength{\tabcolsep}{3.5pt}
\begin{tabular}{lccccc}
\toprule
Method 
& \makecell{\# Params \\ (M)}
& \makecell{MACs \\ (G)}
& \makecell{Time \\ (s)}
& \makecell{Mem. \\ $1024^2$ (GB)}
& \makecell{Mem. \\ $3000{\times}4000$ (GB)}
\\
\midrule
SA-LUT~\cite{salut} 
& 16.68 & 268.16$^\dagger$ & 0.230 & 36.76 & 37.00 \\
Neural Preset~\cite{neuralpreset} 
& 4.66 & 1.13 & 0.016 & 0.35 & 3.69 \\
PhotoWCT2~\cite{photowct2} 
& 7.01 & 759.83$^\ddagger$ & 0.215 & 9.41 & 43.61 \\
\midrule
CDDFuse~\cite{cddfuse} 
& 1.19 & 1872.64 & 0.801 & 6.04 & Failed \\
CDDFuse + flow + photo. 
& 7.55 & 2173.03 & 1.094 & 6.11 & Failed \\
SwinFusion~\cite{swinfusion} 
& 0.98 & 1021.11 & 2.137 & 7.55 & Failed \\
SwinFusion + flow + photo. 
& 7.33 & 1321.50 & 2.339 & 7.67 & Failed \\
\midrule
Ours 
& 7.43 & 812.58 & 0.323 & 2.04 & 22.20$^\ast$ \\
\bottomrule
\end{tabular}

\vspace{0.5em}
\footnotesize{
$^\dagger$SA-LUT MACs are reported at its $512 \times 512$ network input resolution. 
$^\ddagger$PhotoWCT2 MACs do not include the ZCA/SVD stylization core. 
$^\ast$Measured at $3008 \times 4000$ due to padding/alignment constraints.
}
\end{table}

\subsection{Prompt Effects on Hallucination}
\label{appendix:prompts}
We perform an experiment to demonstrate how the hallucination map can be used to analyze which prompts induce fewer hallucinations.
We consider five representative prompts with varying degrees of constraints on structural changes.
Using 20 scenes from the MIT-Adobe FiveK~\cite{ma5k} dataset, we generate two GenAI outputs per scene for each prompt, resulting in 40 input--GenAI pairs per prompt.
For each pair, we obtain the hallucination map $\hat{\mathbf{m}}$ from MSCU-Net and compute a hallucination score as the spatial mean of $\hat{\mathbf{m}}$, averaged over all 40 pairs.
Table~\ref{tab:prompts} reports the prompts and their average hallucination scores.
Visual examples are shown in Figure~\ref{fig:halluc_map_for_prompt}.

 \begin{table}[t]
  \centering                   
  \caption{
    Prompt sensitivity analysis using hallucination maps.
    Lower hallucination scores indicate fewer input--GenAI content mismatches.
  }                                                                                                                                                                                    
  \small                                                                                                                                                                                                              
  \setlength{\tabcolsep}{6pt}                                                                                                                                                                                         
  \begin{tabular}{clp{6.5cm}c}
  \toprule
  \textbf{ID} & \textbf{Type} & \textbf{Prompt} & \textbf{Halluc. Score}$\downarrow$ \\
  \midrule
  0 & Generic, no constraint & ``Make the image more vivid and aesthetically pleasing.'' & 0.0828 \\
  1 & Generic, mild constraint & ``Make the image more vivid and aesthetically pleasing. Preserve the original scene content.'' & 0.0671 \\
  2 & Generic, strong constraint & ``Make the image more vivid and aesthetically pleasing. Strictly preserve scene structure and content. Do not introduce any new objects, textures, or details. Do not modify geometry or spatial layout.'' & 0.0536 \\
  3 & Specific, mild constraint & ``Significantly enhance the image's visual appeal to produce a detail-rich and natural-looking image that appears as if the same scene was captured with a high-end camera, professional lighting, and photofinishing. Preserve the original scene content.'' & 0.0580 \\
  4 & Specific wording & ``Improve color, contrast, and tonal balance to produce a natural, aesthetically pleasing image.'' & 0.0453 \\
  \bottomrule
  \end{tabular}                                                                                                                                                                                                     
  \label{tab:prompts}                                                                                                                                                                                                 
  \end{table}

The results suggest that adding structural constraints to the prompt can reduce hallucinations in the generated images.
However, hallucinations can still occur even with strong structural constraints, as shown in row 2 of Figure~\ref{fig:halluc_map_for_prompt}.
Prompt specificity also appears to matter: 
Prompt 4, which focuses explicitly on color, contrast, and tonal balance, yields the lowest hallucination score.

\subsection{Limitations}
\label{appendix:limitations}
As discussed in Sec.~\ref{sec:conclusion}, 
our framework targets enhancement-based tasks where the camera input serves as a reliable structural reference.
It currently does not address settings where this assumption may not hold, 
such as super-resolution or deblurring; we leave these extensions to future work.

Additionally, our framework mitigates hallucinations, but does not inherently eliminate them.
The central difficulty is distinguishing desired texture enhancement from undesired content changes, 
analogous to the ambiguity between noise and fine texture in image denoising.
Models with sufficient capacity to modify local texture can reduce noise and sharpen edges, 
but may also inadvertently propagate hallucinated content; 
this applies to both our fusion backbone and typical image fusion methods.
LUT-based methods such as SA-LUT~\cite{salut}, which avoid local neighborhood aggregation, 
better preserve input content but have limited ability to denoise or enhance fine texture.

\subsection{Broader Impact}
\label{appendix:impact}
Our work proposes a post-processing framework that fuses a camera image               
with a GenAI-enhanced version to reduce spatial misalignment and content hallucination,                      
targeting enhancement-based tasks such as tone manipulation and low-light enhancement.                       
Our work can enable non-expert users to leverage industrial-grade GenAI tools         
for image retouching while maintaining pixel-level fidelity. At the same time, it provides an interpretable                  
hallucination map that can help users and developers identify and compare the structural                     
faithfulness of different GenAI models and prompts.    


\clearpage
\section*{NeurIPS Paper Checklist}

The checklist is designed to encourage best practices for responsible machine learning research, addressing issues of reproducibility, transparency, research ethics, and societal impact. Do not remove the checklist: {\bf The papers not including the checklist will be desk rejected.} The checklist should follow the references and follow the (optional) supplemental material.  The checklist does NOT count towards the page
limit. 

Please read the checklist guidelines carefully for information on how to answer these questions. For each question in the checklist:
\begin{itemize}
    \item You should answer \answerYes{}, \answerNo{}, or \answerNA{}.
    \item \answerNA{} means either that the question is Not Applicable for that particular paper or the relevant information is Not Available.
    \item Please provide a short (1--2 sentence) justification right after your answer (even for \answerNA). 
\end{itemize}

{\bf The checklist answers are an integral part of your paper submission.} They are visible to the reviewers, area chairs, senior area chairs, and ethics reviewers. You will also be asked to include it (after eventual revisions) with the final version of your paper, and its final version will be published with the paper.

The reviewers of your paper will be asked to use the checklist as one of the factors in their evaluation. While \answerYes{} is generally preferable to \answerNo{}, it is perfectly acceptable to answer \answerNo{} provided a proper justification is given (e.g., error bars are not reported because it would be too computationally expensive'' or ``we were unable to find the license for the dataset we used''). In general, answering \answerNo{} or \answerNA{} is not grounds for rejection. While the questions are phrased in a binary way, we acknowledge that the true answer is often more nuanced, so please just use your best judgment and write a justification to elaborate. All supporting evidence can appear either in the main paper or the supplemental material, provided in appendix. If you answer \answerYes{} to a question, in the justification please point to the section(s) where related material for the question can be found.

IMPORTANT, please:
\begin{itemize}
    \item {\bf Delete this instruction block, but keep the section heading ``NeurIPS Paper Checklist"},
    \item  {\bf Keep the checklist subsection headings, questions/answers and guidelines below.}
    \item {\bf Do not modify the questions and only use the provided macros for your answers}.
\end{itemize}


\begin{enumerate}

\item {\bf Claims}
    \item[] Question: Do the main claims made in the abstract and introduction accurately reflect the paper's contributions and scope?
    \item[] Answer: \answerYes{} 
    \item[] Justification: In the abstract and introduction, we describe the scope of the paper (limited to enhancement-based tasks), and contributions made within this scope.

\item {\bf Limitations}
    \item[] Question: Does the paper discuss the limitations of the work performed by the authors?
    \item[] Answer: \answerYes{} 
    \item[] Justification: We discuss the limitations in Section~\ref{sec:conclusion} and the Appendix~\ref{appendix:limitations}.

\item {\bf Theory assumptions and proofs}
    \item[] Question: For each theoretical result, does the paper provide the full set of assumptions and a complete (and correct) proof?
    \item[] Answer: \answerNA{} 
    \item[] Justification: This paper does not include theoretical results. Equations are used to describe the proposed modules, not to state or prove novel theorems.

    \item {\bf Experimental result reproducibility}
    \item[] Question: Does the paper fully disclose all the information needed to reproduce the main experimental results of the paper to the extent that it affects the main claims and/or conclusions of the paper (regardless of whether the code and data are provided or not)?
    \item[] Answer: \answerYes{} 
    \item[] Justification: We provide a comprehensive description of our experimental setup in Section~\ref{sec:experiments}, which covers datasets, baselines, evaluation metrics, and implementation details. More implementation details are provided in the Appendix.

\item {\bf Open access to data and code}
    \item[] Question: Does the paper provide open access to the data and code, with sufficient instructions to faithfully reproduce the main experimental results, as described in supplemental material?
    \item[] Answer: \answerNo{} 
    \item[] Justification: We currently do not provide code as part of the submission. However, we will release our code and evaluation dataset upon acceptance

\item {\bf Experimental setting/details}
    \item[] Question: Does the paper specify all the training and test details (e.g., data splits, hyperparameters, how they were chosen, type of optimizer) necessary to understand the results?
    \item[] Answer: \answerYes{} 
    \item[] Justification: We provide training and test details in Sec.~\ref{sec:experiments} and Appendix~\ref{appendix:experiments}, including data splits, hyperparameters, and type of optimizer.

\item {\bf Experiment statistical significance}
    \item[] Question: Does the paper report error bars suitably and correctly defined or other appropriate information about the statistical significance of the experiments?
    \item[] Answer: \answerNo{} 
    \item[] Justification: Error bars are not reported as our evaluation follows standard practice in image restoration and enhancement, where results are reported as averages over a fixed test set.

\item {\bf Experiments compute resources}
    \item[] Question: For each experiment, does the paper provide sufficient information on the computer resources (type of compute workers, memory, time of execution) needed to reproduce the experiments?
    \item[] Answer: \answerYes{} 
    \item[] Justification: We report the model complexities and inference cost in Appendix~\ref{appendix:complexity}. All models are trained and evaluated using a single NVIDIA L40 GPU with 48 GB of memory.
    
\item {\bf Code of ethics}
    \item[] Question: Does the research conducted in the paper conform, in every respect, with the NeurIPS Code of Ethics \url{https://neurips.cc/public/EthicsGuidelines}?
    \item[] Answer: \answerYes{} 
    \item[] Justification:  The methods and datasets presented in this paper comply with the NeurIPS Code of Ethics.

\item {\bf Broader impacts}
    \item[] Question: Does the paper discuss both potential positive societal impacts and negative societal impacts of the work performed?
    \item[] Answer: \answerYes{} 
    \item[] Justification: We discuss the potential positive societal impacts in Appendix~\ref{appendix:impact}. 
    Our method is a post-processing framework for color and tonal alignment that                                 
    preserves the structural content of the original image; we do not foresee                                    
    significant negative societal impacts.  
    
\item {\bf Safeguards}
    \item[] Question: Does the paper describe safeguards that have been put in place for responsible release of data or models that have a high risk for misuse (e.g., pre-trained language models, image generators, or scraped datasets)?
    \item[] Answer: \answerNA{} 
    \item[] Justification: The paper does not pose such risks.

\item {\bf Licenses for existing assets}
    \item[] Question: Are the creators or original owners of assets (e.g., code, data, models), used in the paper, properly credited and are the license and terms of use explicitly mentioned and properly respected?
    \item[] Answer: \answerYes{} 
    \item[] Justification: All datasets and models used are properly cited.                                      
  The MIT-Adobe FiveK dataset is released for academic research use;                                           
  LOL v1, LOL v2, and SICE are publicly available for non-commercial research.                                 
  Pretrained models (SKFlow, NAFNet) are used under their respective open-source licenses.                     
  We will provide full license details upon code release. 

\item {\bf New assets}
    \item[] Question: Are new assets introduced in the paper well documented and is the documentation provided alongside the assets?
    \item[] Answer: \answerNA{} 
    \item[] Justification: No new assets are released with this submission.

\item {\bf Crowdsourcing and research with human subjects}
    \item[] Question: For crowdsourcing experiments and research with human subjects, does the paper include the full text of instructions given to participants and screenshots, if applicable, as well as details about compensation (if any)? 
    \item[] Answer: \answerNA{} 
    \item[] Justification: This paper does not involve human subjects.

\item {\bf Institutional review board (IRB) approvals or equivalent for research with human subjects}
    \item[] Question: Does the paper describe potential risks incurred by study participants, whether such risks were disclosed to the subjects, and whether Institutional Review Board (IRB) approvals (or an equivalent approval/review based on the requirements of your country or institution) were obtained?
    \item[] Answer: \answerNA{} 
    \item[] Justification: This paper does not involve human subjects.

\item {\bf Declaration of LLM usage}
    \item[] Question: Does the paper describe the usage of LLMs if it is an important, original, or non-standard component of the core methods in this research? Note that if the LLM is used only for writing, editing, or formatting purposes and does \emph{not} impact the core methodology, scientific rigor, or originality of the research, declaration is not required.
    \item[] Answer: \answerNA{} 
    \item[] Justification: The core method development in this research does not involve LLMs as any important, original, or non-standard components

\end{enumerate}

\end{document}